\documentclass[10pt,twocolumn,letterpaper]{article}

\usepackage[pagenumbers]{cvpr} %

\usepackage{graphicx}
\usepackage{amsmath}
\usepackage{amssymb}
\usepackage{booktabs}
\usepackage{comment}
\usepackage{multirow}
\usepackage{tabularx}
\usepackage{float}
\usepackage{xcolor}

\usepackage{import}
\usepackage{comment}

\graphicspath{{figures/}}

\usepackage[pagebackref,breaklinks,colorlinks]{hyperref}

\usepackage[capitalize]{cleveref}
\crefname{section}{Sec.}{Secs.}
\Crefname{section}{Section}{Sections}
\Crefname{table}{Table}{Tables}
\crefname{table}{Tab.}{Tabs.}

\newcommand\blfootnote[1]{%
  \begingroup
  \renewcommand\thefootnote{}\footnote{#1}%
  \addtocounter{footnote}{-1}%
  \endgroup
}

\begin{document}

\newcommand\appx[1] {#1}

\title{Neural Head Avatars from Monocular RGB Videos}

\author{
Philip-William Grassal$^{1*}$
\and
Malte Prinzler$^{1*}$
\and
Titus Leistner$^1$
\and
Carsten Rother$^1$\vspace{0.1cm}
\and
Matthias Nießner$^2$
\and
Justus Thies$^3$
\and
\\
$^1$Heidelberg University~~~~
$^2$Technical University of Munich~~~~
$^3$Max Planck Institute for Intelligent Systems
}

\twocolumn[{
	\renewcommand\twocolumn[1][]{#1}%
	\maketitle
 	\begin{center}
 	\def\svgwidth{\linewidth}
    \begingroup%
  \makeatletter%
  \providecommand\color[2][]{%
    \errmessage{(Inkscape) Color is used for the text in Inkscape, but the package 'color.sty' is not loaded}%
    \renewcommand\color[2][]{}%
  }%
  \providecommand\transparent[1]{%
    \errmessage{(Inkscape) Transparency is used (non-zero) for the text in Inkscape, but the package 'transparent.sty' is not loaded}%
    \renewcommand\transparent[1]{}%
  }%
  \providecommand\rotatebox[2]{#2}%
  \newcommand*\fsize{\dimexpr\f@size pt\relax}%
  \newcommand*\lineheight[1]{\fontsize{\fsize}{#1\fsize}\selectfont}%
  \ifx\svgwidth\undefined%
    \setlength{\unitlength}{1228.00391971bp}%
    \ifx\svgscale\undefined%
      \relax%
    \else%
      \setlength{\unitlength}{\unitlength * \real{\svgscale}}%
    \fi%
  \else%
    \setlength{\unitlength}{\svgwidth}%
  \fi%
  \global\let\svgwidth\undefined%
  \global\let\svgscale\undefined%
  \makeatother%
  \begin{picture}(1,0.22470208)%
    \lineheight{1}%
    \setlength\tabcolsep{0pt}%
    \put(0,0){\includegraphics[width=\unitlength,page=1]{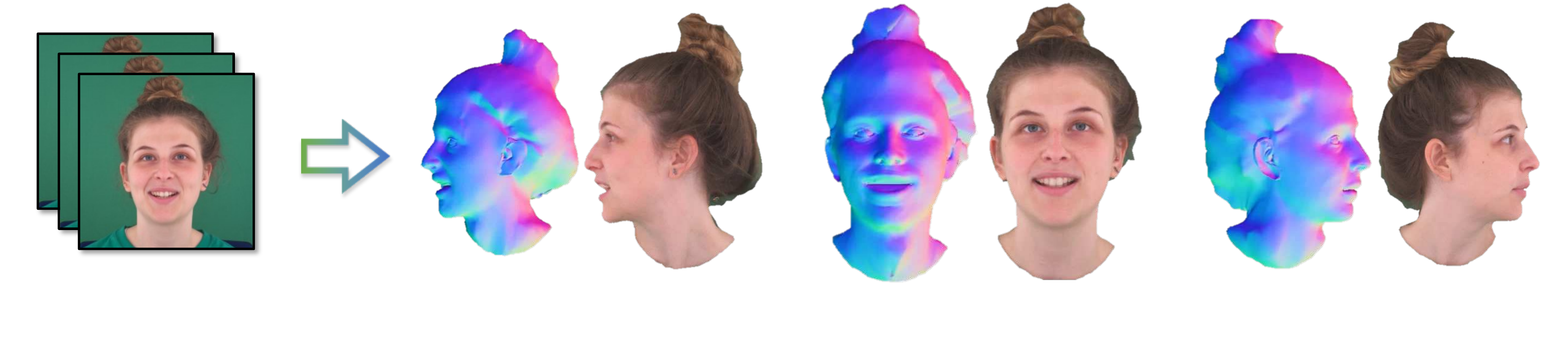}}%
    \put(0.00274376,0.01663275){\color[rgb]{0,0,0}\makebox(0,0)[lt]{\lineheight{1.25}\smash{\begin{tabular}[t]{l}Monocular RGB Video\end{tabular}}}}%
    \put(0.33144053,0.01967759){\color[rgb]{0,0,0}\makebox(0,0)[lt]{\lineheight{1.25}\smash{\begin{tabular}[t]{l}Neural Head Avatar with Articulated Geometry \& Photorealistic  Texture\end{tabular}}}}%
  \end{picture}%
\endgroup%

		\captionof{figure}{
		    Given a monocular portrait video of a person, we reconstruct a \textit{Neural Head Avatar}.
		    Such a 4D avatar will be the foundation of applications like teleconferencing in VR/AR, since it enables novel-view synthesis and control over pose and expression.
		}
		\label{fig:teaser}
	\end{center}
}]

\blfootnote{* Both authors contributed equally to the paper}

\begin{abstract}
We present \textit{Neural Head Avatars}, a novel neural representation that explicitly models the surface geometry and appearance of an animatable human avatar that can be used for teleconferencing in AR/VR or other applications in the movie or games industry that rely on a digital human.\footnote{ \tiny \url{philgras.github.io/neural_head_avatars/neural_head_avatars.html}}
Our representation can be learned from a monocular RGB portrait video that features a range of different expressions and views.
Specifically, we propose a hybrid representation consisting of a morphable model for the coarse shape and expressions of the face, and two feed-forward networks, predicting vertex offsets of the underlying mesh as well as a view- and expression-dependent texture.
We demonstrate that this representation is able to accurately extrapolate to unseen poses and view points, and generates natural expressions while providing sharp texture details. 
Compared to previous works on head avatars, our method provides a disentangled shape and appearance model of the complete human head (including hair) that is compatible with the standard graphics pipeline.
Moreover, it quantitatively and qualitatively outperforms current state of the art in terms of reconstruction quality and novel-view synthesis.
\end{abstract}

\section{Introduction}
Reconstructing and reenacting human heads has been a long studied research problem and will be a key driver for future applications in VR/AR, teleconferencing, games and the movie industry.
For those applications, it is of particular interest to get an accurate 3D shape and appearance model that provides 3D consistency and strong identity preservation under novel view points, poses and expressions.
Reconstructing such a model, especially from monocular input data (e.g., from a webcam), is difficult due to the complex geometry of facial dynamics and the missing 3D information~\cite{Zollhoefer2018FaceSTAR}. 
Indeed, several state-of-the-art methods for talking head synthesis avoid explicit geometry reconstruction and rely on image or feature-based warping for motion control and generative networks for image synthesis \cite{nvidia, FOMM, bilayer}.
These methods are generalized and deliver impressive reenactment results even with only a single input image of the subject.
However, the quality of the approaches drops significantly for larger changes in pose or view point as no 3D-consistent geometry representation is used.
Shape proxies such as a 3D morphable model~\cite{3dmm,3DMM_survey} can be utilized to improve the 3D consistency of synthetic faces~\cite{face2face, deferredneuralrendering, varitex, deepvideoportraits} since the facial information is embedded on the proxy surface.
Besides image- or surface-based representations, volumetric representations are used~\cite{neural_volumes, nerface}.
While they show promising results without an explicit surface prior, these methods still lack a consistent full head shape reconstruction from single view inputs and are not compatible with the standard rasterization pipelines.

In this work, we present \textit{Neural Head Avatars}, an explicit shape and appearance representation of the complete human head (including hair) which can be used in existing graphics pipelines that use triangular meshes.
Specifically, we employ coordinate-based multi-layer perceptrons (MLPs) to predict the 3D meshes and dynamic textures, depending on the facial expression and pose of real humans.
These networks are embedded on the surface of the FLAME morphable head model~\cite{flame} which also serves as a coarse shape and expression proxy.
We show that one can optimize such an explicit head representation based on a short monocular RGB video sequence.
Using color-dependent and color-independent energy terms during optimization, we disentangle the reconstruction of surface geometry and color detail.
The resulting controllable 4D avatar (3D model + motion) is subject-specific and generates novel poses and expressions while preserving high photo-realism.
Moreover, it demonstrates great visual quality under large view point changes and, therefore, addresses one of the main drawbacks of related approaches.

\smallskip\smallskip
\noindent
In summary, our contributions are:
\vspace{-0.12cm}
\begin{itemize}
  \setlength{\itemsep}{2pt}
  \setlength{\parskip}{2pt}
  \setlength{\parsep}{2pt}
	\item Neural Head Avatars, a novel, subject-specific representation for articulated human heads that explicitly reconstructs the full head geometry and produces photo-realistic results even under large view point changes, 
	\item A fully differentiable optimization pipeline to optimize Neural Head Avatars from a short, monocular RGB video with color-dependent and color-independent energy terms that allow for the disentanglement of the surface shape and color detail. 
\end{itemize}

\begin{figure*}[h]
\includegraphics[width=\textwidth]{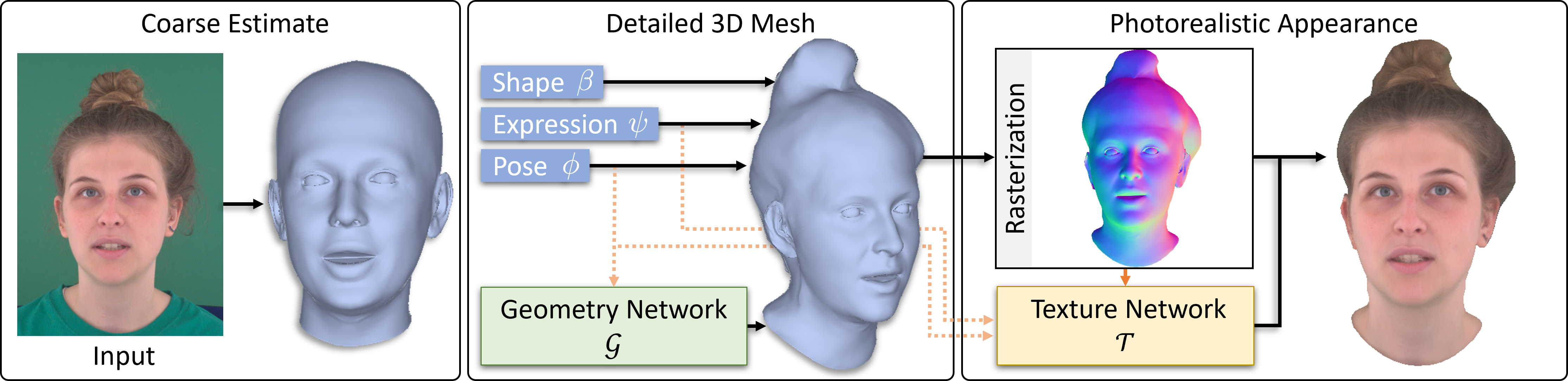}
\caption{Overview of our method.
Given an RGB input frame, we use an adaption of the real-time face tracker \cite{face2face} to estimate low-dimensional shape, expression, and pose parameters of the linear head model FLAME \cite{flame}.
We adapt the geometry generated by FLAME with a geometry refinement network $\mathcal{G}$. The resulting mesh is rasterized with a standard computer graphics pipeline.
The texture network $\mathcal{T}$ synthesizes the mesh texture of the rasterized surface. Conditioning $\mathcal{T}$ on the canonical surface position, a local normal patch, and flame parameters enables the synthesis of view- and expression-dependent effects.
The networks and FLAME parameters are optimized in an analysis-by-synthesis fashion  with color-dependent and color-independent energy terms that allow for the disentanglement of the surface shape and color detail.
The reconstructed \textit{Neural Head Avatar} can be animated using the expression and pose parameters $\psi, \theta$, and can be rendered under novel viewpoints.
}
\label{fig:method}
\end{figure*}

\section{Related Work}
Reconstructing controllable 4D head or facial avatars is an actively studied field at the intersection of computer vision and computer graphics.
For an extensive review of methods, we refer to the state of the art report of Zollhöfer et al.~\cite{Zollhoefer2018FaceSTAR}.

\paragraph{Image-based models.}
Image-based models synthesize the face of a subject without relying on any (explicit or implicit) representation in 3D space.
These methods either utilize (learned) warping fields \cite{FOMM,elor2017bringingPortraits} to deform an input image to match new poses or expressions, or deploy encoder-decoder architectures, where the encoder extracts an identity code from a given source image and a decoder synthesizes the output image \cite{bilayer,fsgan,pix2pix_contour,oneshotfacereenactment}.
The decoder may be conditioned by facial landmarks \cite{bilayer,fsgan}, facial contours \cite{pix2pix_contour}, or parsing maps \cite{oneshotfacereenactment}.
Even though these methods produce high-quality results and even allow for real-time synthesis \cite{bilayer}, they suffer from artifacts for strong pose and expression changes, and lack geometric and temporal consistency.
This is mainly due to the fact, that the appearance of deformations in three-dimensional space (e.g., yaw opening, head rotation) must be learned in 2D by these models.

\paragraph{Implicit models.}
Implicit models represent the geometry using implicit surface functions (e.g., signed distance functions) or by volumetric representations.
A common approach is to represent the appearance of a target person in a discrete latent feature voxel grid that can be deformed to synthesize dynamic deformations \cite{nvidia, neural_volumes}.
Motivated by their recent success in 3D scene reconstruction~\cite{tewari2021advances}, neural radiance fields (NeRF) in combination with volumetric rendering \cite{nerf} have been used to replace the discrete feature voxel grids \cite{pixelaligned_volumetric_avatars, nerface, neural_lumigraph_rendering, park2021nerfies,park2021hypernerf,Noguchi2021,su2021anerf,peng2021animatable,Peng_2021_CVPR,liu2021neural,Xu2021,Alldieck_2021_ICCV}.
Articulated head avatars can be synthesized by conditioning the NeRF on low-dimensional parameters of a face model \cite{nerface,wang2020learning} or audio signals~\cite{guo2021adnerf}.
Pixel-aligned Volumetric avatars~\cite{pixelaligned_volumetric_avatars} are generalized across subjects  and can generate novel views, based on single or multiple input images.
Even though solving geometric and temporal inconsistencies, the proposed methods either fail to disentangle pose and expression \cite{nvidia}, are limited to static reconstructions \cite{h3dnet,neural_lumigraph_rendering, pixelaligned_volumetric_avatars, WangPrior2021} or fail to generalize to unseen poses and expressions \cite{nerface}.

\paragraph{Explicit models.}
The majority of head reconstruction methods relies on explicit scene representations, i.e., triangular meshes~\cite{Zollhoefer2018FaceSTAR,Garrido2013,Garrido2014,Garrido2015,Weise2009,Weise2011,Thies15,face2face,Thies_2018,Blanz2003,Blanz2004,deepvideoportraits,deferredneuralrendering,thies2020nvp,Chen:2013,Li2013,Ichim2015,Bouaziz:2013,hsieh2015unconstrained,chen2013,chen_2014}.
For these methods, morphable models are used as a prior to reconstruct the face from incomplete (e.g., partially occluded) or noisy data (e.g., from depth maps).
Morphable models are computed from a population of 3D head scans~\cite{3DMM_survey}, and provide statistical information on physiologically plausible head shapes and facial movements \cite{bfm, flame, lyhm, lsfm}.
In addition to the geometry, these models can provide a statistical linear model for the texture \cite{3dmm, lineartexturemodel, bfm, flametexture, 3dmm_facerec} which can be used to reconstruct faces from RGB data only~\cite{Zollhoefer2018FaceSTAR,3dmm,face2face}.
Recent work utilizes generative adversarial networks (GAN) \cite{gan,pagan} to generate and optimize albedo and normal maps for specific subjects \cite{ganfit,avatarme}.
Other approaches utilize 2D neural rendering~\cite{tewari2020neuralrendering,tewari2021advances}, to learn how to render photo-realistic imagery of a specific subject from a short training dataset~\cite{deepvideoportraits,deferredneuralrendering,thies2020nvp}.
These approaches are based on deep neural networks which can be conditioned on coarse RGB renderings based on a linear texture model \cite{deepvideoportraits}, uv-maps \cite{varitex}, latent feature maps \cite{deferredneuralrendering} or point clouds \cite{smplpix}.
While these methods produce geometrically consistent avatars that can be easily controlled, they either are limited to craniofacial structures and do not include the synthesis of hair \cite{ganfit, avatarme} or suffer from temporal and spatial inconsistencies due to their loose bound to the geometric backbone \cite{varitex, deferredneuralrendering, smplpix}.
Other methods model mesh displacements to reconstruct fine detail wrinkles and hair \cite{detailedhumanavatars, deca, chen2015}, therefore, maintaining a close bound to the underlying geometry. However, none of them produces photo-realistic outputs under novel views.

In contrast to previous work, we jointly optimize a photo-realistic dynamic texture together with subject specific geometry that includes facial detail and hair structure, stored in a coordinate-based, fully connected neural network.
By deploying a texture-based approach, we tightly link appearance and underlying geometry which ensures spatial consistency and generalization to unseen poses and expressions.
Articulated jaw, neck, and eyes allow for intuitive avatar control.
We demonstrate that an avatar can be optimized from a short monocular RGB sequence without the need of special (multi-view) camera setups as in \cite{PiCA, deepappearance}.

\section{Method}
\label{sec:method}

Given an RGB video sequence of a talking person consisting of $N$ consecutive frames $I_1, I_2, ..., I_N$, we reconstruct a 4D neural avatar based on an explicit representation that allows for pose- and expression-dependent novel viewpoint synthesis.
Specifically, our model outputs a classical triangle mesh, i.e., vertices $V = (v_1, v_2, ..., v_n), v_i \in \mathbb{R}^3$, connecting faces $F$ and a texture function $\mathcal{T}$ that assigns an RGB color value to each point on the surface defined by $V$ and $F$ (see \Cref{sec:method:components}).
Thus, the standard graphics pipeline can be utilized to obtain a rendered image $\hat{I}$ assuming a full perspective camera projection.
Based on this image formation model, we optimize our avatar representation in an analysis-by-synthesis-based fashion (see \Cref{sec:method:optimization}).
An overview of our method is depicted in \Cref{fig:method}.

\subsection{Explicit Neural Head Representation}
\label{sec:method:components}
Our explicit surface representation is embedded on the FLAME~\cite{flame} template surface and shares its topology $F$.
Specifically, we employ a multi-layer perceptron (MLP) $\mathcal{G}$ which models the pose-dependent offsets w.r.t. the template surface.
To generate the view-, pose-, and expression-dependent texture of the face, we use an MLP $\mathcal{T}$ which predicts a color value at any surface point of the mesh.

\paragraph{Template Model.}
We deploy the parametric FLAME head model \cite{flame} as a geometric backbone of our method:
\begin{equation*}
    \begin{split}
        V_\text{flame}: \mathbb{R}^{300}, \mathbb{R}^{100}, \mathbb{R}^{3k} &\to \mathbb{R}^{16227\times 3} \\
        \beta, \psi, \phi &\mapsto V_\text{flame}(\beta, \psi, \phi)
    \end{split}
    \label{eq:flame}
\end{equation*}
where $\beta$, $\psi$ and $\phi$ describe shape, expression and $k=4$ joint pose parameters, respectively.
We perform minor adjustments to the FLAME topology, namely, we uniformly subdivide the faces (four-way subdivision), remove the faces belonging to the lower neck region and add additional faces to close the mouth cavity.
This increases the original number of vertices from 5023 to 16227.

\paragraph{Geometry Refinement Network $\mathcal{G}$.}
To model facial detail and hair which is not represented by the FLAME head model, we introduce a pose-dependent offset function for geometry corrections:
\begin{equation*}
    \begin{split}
        \mathcal{G}: \mathbb{R}^{3k} &\to \mathbb{R}^{16227\times 3} \\
        \phi &\mapsto \mathcal{G}(\phi).
    \end{split}
    \label{eq:flame}
\end{equation*}
Using this offset function, the mesh geometry is given by:%
\begin{equation*}
    V(\beta, \psi, \phi) = V_\text{flame}(\beta,\psi, \phi) + \mathcal{G}(\phi).
\end{equation*}

\paragraph{Texture Network $\mathcal{T}$.}
While \cite{flametexture} provides a linear texture space for the FLAME head model, due to its Gaussian nature it lacks fine detail and photo-realism.
We introduce a novel appearance model $\mathcal{T}$ which generates a photo-realistic texture, including the synthesis of expression and view dependent effects.
In order to predict the color of a point on the mesh, $\mathcal{T}$ receives the 3D coordinates of that point on the canonical FLAME template mesh, the expression and pose of the current frame, and a local patch of the rendered normals as input, and returns the estimated color value.
This conditioning of $\mathcal{T}$ enables the synthesis of expression- and view-dependent effects (compare \Cref{fig:ablations} (b)).
Formally, $\mathcal{T}$ performs the mapping:
\begin{equation*}
    \begin{split}
        \mathcal{T}: \mathbb{R}^{3}, \mathbb{R}^{100}, \mathbb{R}^{3k}, \mathbb{R}^{n\times n\times 3} &\to \mathbb{R}^{3} \\
        p_i, \psi, \phi, \hat{N}_i &\mapsto c_i
    \end{split}
\end{equation*}
with $c_i$ denoting the predicted color at pixel $i$, $\hat{N}_i$ a local patch of the rendered normal map around $i$ and $p_i$ being the 3D location on the canonical FLAME template mesh depicted in $i$.

\begin{figure}
     \centering
     \begin{subfigure}[b]{0.3\linewidth}
         \centering
         \includegraphics[width=\textwidth]{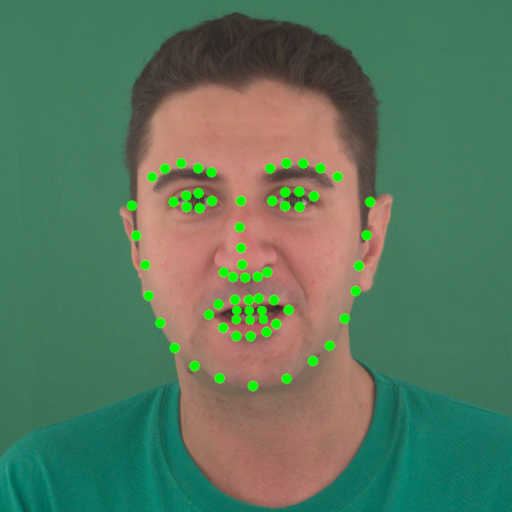}
         \caption{Landmarks~\cite{facealignment}}
     \end{subfigure}
     \hfill
     \begin{subfigure}[b]{0.3\linewidth}
         \centering
         \includegraphics[width=\textwidth]{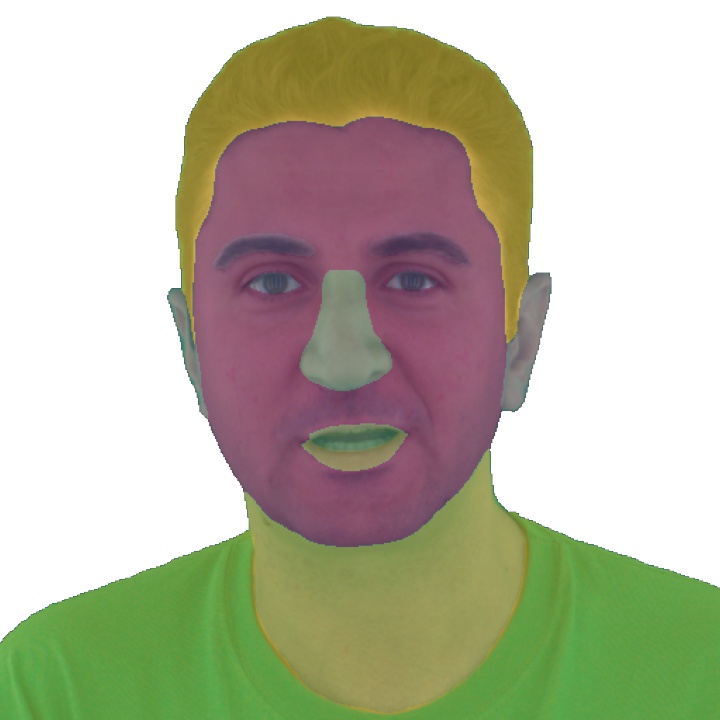}
         \caption{Semantic Labels~\cite{faceparsing}}
     \end{subfigure}
     \hfill
     \begin{subfigure}[b]{0.3\linewidth}
         \centering
         \includegraphics[width=\textwidth]{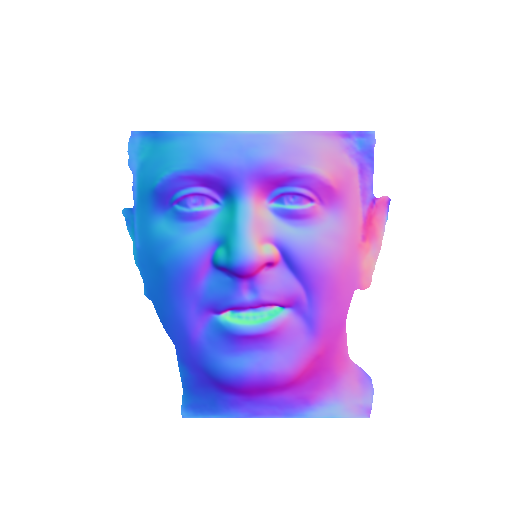}
         \caption{Normals~\cite{facenormals}}
     \end{subfigure}
        \caption{Our optimization is based on the color data of a short video sequence, the corresponding detected facial landmarks (a), predicted semantic labels (b) and predicted normal maps (c). }
        \label{fig:inputs}
\end{figure}

\medskip
We approximate both functions $\mathcal{G}$ and $\mathcal{T}$ using two subject-specific, coordinate-based multi-layer perceptrons~\cite{piGAN}.
\appx{Please refer to \Cref{appx:network-architectures} for additional implementation details.}

\subsection{Optimization based on Monocular RGB Data}
\label{sec:method:optimization}

The joint optimization of head geometry and texture is a highly underconstrained optimization problem for short monocular video sequences.
Besides the data terms that are based on the RGB inputs%
, we employ regularization strategies that ensure smooth reconstructed surfaces, and view-consistent texture synthesis.
The objective of the joint optimization $E_\text{joint}$ is defined as:
\begin{equation}
    \begin{split}
        E_\text{joint} = E_\text{geom} + E_\text{app},
    \end{split}
    \label{eq:objective}
\end{equation}
where $E_\text{geom}$ measures the data and regularization terms w.r.t. the geometry and $E_\text{app}$ contains the terms w.r.t. the appearance, i.~e., texture and color reproduction.

\paragraph{Geometry Objective $E_\text{geom}$.}
To disentangle appearance and geometry, we define a geometry energy term which is independent of the actual appearance:
\begin{equation}
    \begin{split}
    E_\text{geom} &=  w_\text{lmk} \cdot E_\text{lmk} + w_\text{normal} \cdot E_\text{normal}\\
                    &+w_\text{semantic} \cdot E_\text{semantic} +w_\text{reg,geom} \cdot E_\text{reg, geom}. \\
    \end{split}
    \label{eq:engergies:geom}
\end{equation}

The landmark energy $E_\text{lmk}$ measures the $\ell_1$ distance of detected 2D facial landmarks~\cite{facealignment,mediapipe} and the projected counterparts on the mesh surface.
Besides the absolute positions of the landmarks, it also measures the relative distances of the eye landmarks at the upper and lower lid to improve the reconstruction of eye lid closure~\cite{deca} (see ablation study in \Cref{fig:ablations} (c)).

The energy term $E_\text{normal}$ is based on pseudo-normal maps $N\in \mathbb{R}^{H \times W \times 3}$ using the pretrained model of Abrevaya et al.~\cite{facenormals}.
Based on these predictions, we formulate a reconstruction energy term for fine geometric detail.
As our focus lies on high-frequency geometry detail, instead of minimizing the absolute difference between pseudo-normals $N$ and our predicted normals $\hat{N}$, we instead match their image-space Laplacians $\lambda(\cdot)$:
\begin{equation*}
        E_\text{normal} = |\lambda(\hat{N}) - \lambda(N)|_1 .
\end{equation*}

We employ $E_\text{semantic}$, to match the semantic regions $S_k$ of the input and $\hat{S}_k$ of the reconstructed mesh for \emph{facial skin and neck}, \emph{eyes}, \emph{ears}, and \emph{hair}:
\begin{equation*}
\begin{split}
        E_\text{semantic} =& \sum_{k=1}^{4} S_k \oplus \hat{S}_k,
\end{split}
\end{equation*}
where $\oplus$ denotes an xor on the region.
The respective semantic maps $S_k$ are computed using \cite{faceparsing, bisenet} (see \Cref{fig:inputs} (b)).

Besides the data terms, we employ a regularization term $E_\text{reg, geom}$ which regularizes the FLAME parameters, as well as the geometry MLP $\mathcal{G}$:
\begin{equation*}
E_\text{reg,geom} = w_\text{reg, flame} \cdot E_\text{reg, flame} + w_\text{reg,offset} \cdot E_\text{reg,offset}.
\end{equation*}
Following \cite{3dmm,face2face}, $E_\text{reg,flame}$ uses the statistical properties of the linear shape model, and regularizes the prediction towards the canonical template head using an $\ell_2$-norm on $\beta, \psi, \phi$.
The offsets are regularized using $E_\text{reg,offset}$ which consists of a Laplacian regularizer and a regularizer that controls the pose-consistency and distribution of the predicted offsets.
\appx{We refer to \Cref{appx:optimization-from-monocular-rgb} for additional details.}

\paragraph{Appearance Objective $E_\text{app}$.}
The appearance term $E_\text{app}$ measures the reproduction of the color image $I$.
It depends on both the geometry as well as the texture parameters of our neural head model.
We use dense per-pixel energy terms $E_\text{phot}$, as well as an energy term that measures the perceptual distance $E_\text{perc}$ \cite{perceptual_loss}:
\begin{equation}
    \begin{split}
        E_\text{app} = &~w_\text{phot} \cdot E_\text{phot} + w_\text{perc} \cdot E_\text{perc}.
    \end{split}
    \label{eq:energies:tex}
\end{equation}
$E_\text{perc}$ compares image features of predicted and ground truth images extracted by the face detector from \cite{arcface}.

\paragraph{Initialization and Optimization Strategy}
\label{sec:method:optimization:facetracking}
To initialize our reconstruction method, we adopt the tracking algorithm proposed by \cite{face2face} which optimizes camera, shape, expression, and pose parameters based on an analysis-by-synthesis approach, using the FLAME model~\cite{flame} with a linear texture space~\cite{flametexture}.
The resulting reconstruction is in coarse alignment with the training sequence, but the FLAME model is limited to bald heads and lacks fine, subject-specific geometric detail (see \Cref{fig:method}).
We initialize the geometry refinement network $\mathcal{G}$, by optimizing only for the $E_\text{geom}$ term defined in \Cref{eq:engergies:geom}.
Once, we obtained an estimate of the full head geometry, we optimize for the texture MLP parameters w.r.t. $E_\text{app}$ (\Cref{eq:energies:tex}).
Based on this initialization scheme, we optimize jointly the geometry and texture parameters to minimize $E_\text{joint}$ (\Cref{eq:objective}).
\appx{For implementation details, we refer to \Cref{sec:implementation_details}.\footnote{Additionally, we will release the code for research purposes.}}

\section{Results}
We quantitatively and qualitatively evaluate the performance of our model on the tasks of geometry reconstruction as well as novel pose-, expression-, and view synthesis, and compare it to state-of-the-art methods. 

\subsection{Datasets}
\label{sec:data}
Since the current literature does not provide suitable datasets for the evaluation of dynamic full head approaches, we created two datasets.
\paragraph{Synthetic Data.}
Our synthetic dataset has been generated with the open source \emph{MakeHuman} project \cite{makehuman} which allows to model fully animatable and texturized human models with high variance in appearance and facial geometry.
We generated two female and two male subjects with different ethnicity and head geometry, and render animated sequences (200 training frames, 210 validation frames).
The resulting sequences provide ground truth RGB-, normal- and semantic maps as well as landmarks and 3D meshes which is used to quantitatively evaluate the geometry reconstruction of our method.
Consequently, for evaluation purposes, we will not rely on predicted pseudo-ground truth (normal and semantic maps, landmarks) for experiments with this dataset.
Note that the synthetic meshes have different topologies compared to FLAME and our model.

\paragraph{Real Data.}
Our real dataset contains four sequences of humans, two with male actors, two with a female actor.
The sequences capture various hair styles such as short hair, long hair and a hair bun.
All sequences show the subjects during a natural conversation in front of a green screen in an environment with uniform lighting.
We capture $750$ training frames and $750$ validation frames for each sequence and ensure that both sides of the head are visible at least once in the training partition.
We complement the obtained RGB ground truth with detected facial landmarks as well as normal- and semantic maps by deploying pretrained models \cite{facealignment, facenormals, faceparsing}.
The resulting dataset is used to qualitatively and quantitatively evaluate our model on the task of novel pose-, expression-, and view synthesis, and to compare to state-of-the-art methods.

\subsection{Geometry Reconstruction Quality}
\begin{figure}[t]
    \centering
    \def\svgwidth{\linewidth}
    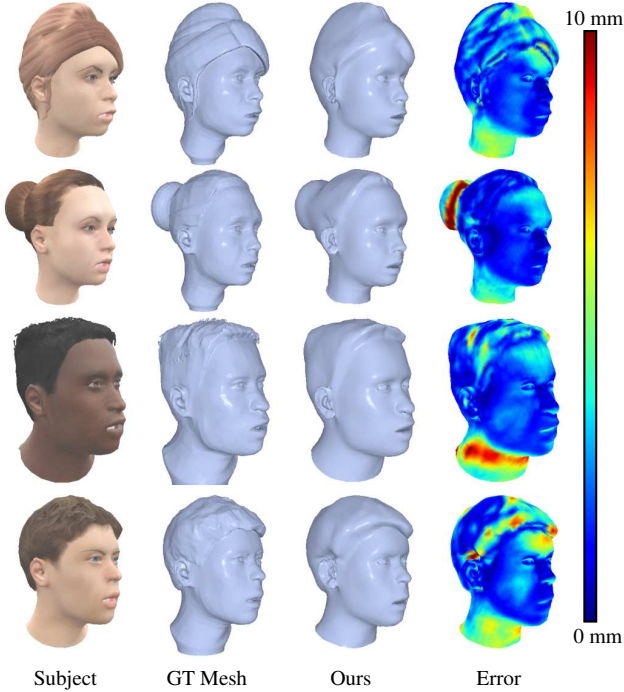
    \caption[Mesh-to-mesh comparison]{From left to right: Ground truth image of the synthetic subject, its ground truth geometry, and our predicted geometry. On the right, we show the average Hausdorff distance in mm (from prediction to the ground-truth mesh) over all validation frames.}
    \label{fig:geometry-evaluation}
\end{figure}

\begin{table}
\resizebox{\linewidth}{!}{%
    \begin{tabular}{lrrrrr} \toprule 
        \textbf{Metric} & \textbf{Female 1} & \textbf{Female 2} & \textbf{Male 1} & \textbf{Male 2}\\ \midrule 
        Normal: FLAME & $15.8^{\circ}$ & $13.3^{\circ}$ & $15.0^{\circ}$ & $14.8^{\circ}$\\
        Normal: Ours & $\mathbf{14.4^{\circ}}$ & $\mathbf{12.2^{\circ}}$ & $\mathbf{13.7^{\circ}}$ & $\mathbf{13.7^{\circ}}$\\
        \midrule 
        Hausdorff: FLAME (Face) & 1.4 & 0.9 & 1.5 & \textbf{1.1}\\
        Hausdorff: Ours (Face) & \textbf{1.2} & \textbf{0.9} & \textbf{1.4} & 1.2\\
        \midrule
        Hausdorff: FLAME & 5.5 & 4.8 & 6.1 & 5.7\\
        Hausdorff: Ours & \textbf{2.6} & \textbf{2.5} & \textbf{3.0} & \textbf{3.1} \\
        \bottomrule 
    \end{tabular}

}
\caption{For four synthetic characters shown in \Cref{fig:geometry-evaluation}, we evaluate our shape prediction using the validation sequence (210 frames) of our dataset.
We list the averaged normal error (angular error) and the average mesh alignment error (Hausdorff distance in mm).
Normal vectors are compared per pixel in the rendered image, masked by the head region. %
The single-sided Hausdorff distance is computed from prediction to ground truth meshes, on either the full head or the facial region only. %
We compare against the reconstruction by FLAME~\cite{flame} as a baseline.
}
\label{tab:geometry-evaluation}
\end{table}
To quantify the head shape reconstruction quality, we utilize the rendered ground truth normals, as well as the meshes provided by the synthetic recordings. 
Those are compared with the predicted meshes and normal maps by computing their single-sided Hausdorff distance in millimeters (mesh alignment error, as in \cite{3dfaw}) and their normal angular error on the validation sequence.
The Hausdorff distance is computed once for the full head and once for the facial region only.
We compare with the reconstructions by FLAME~\cite{flame} obtained from our tracker as a baseline.
The quantitative results are reported in \Cref{tab:geometry-evaluation}.
Our reconstruction and the misalignment error w.r.t. the ground truth are visualized in \Cref{fig:geometry-evaluation}.
We can see that our approach reconstructs the talking head faithfully, and even regions, where only a few silhouette views are available (side of the bun or front/back of the neck) can be estimated, however, with a slightly higher reconstruction error.
\appx{The reconstruction of real subjects in neutral pose is compared with  multi-view stereo (MVS) recordings in \Cref{fig:geometry-evaluation-real} in the Appendix.} %

\subsection{Novel Pose and Expression Synthesis}

\begin{figure*}[t]
    \centering
    \def\svgwidth{\linewidth}
    \begingroup%
  \makeatletter%
  \providecommand\color[2][]{%
    \errmessage{(Inkscape) Color is used for the text in Inkscape, but the package 'color.sty' is not loaded}%
    \renewcommand\color[2][]{}%
  }%
  \providecommand\transparent[1]{%
    \errmessage{(Inkscape) Transparency is used (non-zero) for the text in Inkscape, but the package 'transparent.sty' is not loaded}%
    \renewcommand\transparent[1]{}%
  }%
  \providecommand\rotatebox[2]{#2}%
  \newcommand*\fsize{\dimexpr\f@size pt\relax}%
  \newcommand*\lineheight[1]{\fontsize{\fsize}{#1\fsize}\selectfont}%
  \ifx\svgwidth\undefined%
    \setlength{\unitlength}{3150bp}%
    \ifx\svgscale\undefined%
      \relax%
    \else%
      \setlength{\unitlength}{\unitlength * \real{\svgscale}}%
    \fi%
  \else%
    \setlength{\unitlength}{\svgwidth}%
  \fi%
  \global\let\svgwidth\undefined%
  \global\let\svgscale\undefined%
  \makeatother%
  \begin{picture}(1,0.58630918)%
    \lineheight{1}%
    \setlength\tabcolsep{0pt}%
    \put(0.53249172,0.34867428){\color[rgb]{0,0,0}\makebox(0,0)[lt]{\begin{minipage}{0.05513548\unitlength}\centering \end{minipage}}}%
    \put(0.00010238,0.58749975){\color[rgb]{0,0,0}\makebox(0,0)[lt]{\begin{minipage}{0.99989756\unitlength}\begin{tabularx}{\textwidth}{*7{>{\centering\arraybackslash}X}} VariTex \cite{varitex} & Bi-Layer \cite{bilayer} & FOMM \cite{FOMM} & DVP \cite{deepvideoportraits} & NerFACE \cite{nerface} & Ours & Ground Truth \\ \end{tabularx}\end{minipage}}}%
    \put(0.5456863,0.36676278){\color[rgb]{0,0,0}\makebox(0,0)[lt]{\begin{minipage}{0.11683315\unitlength}\raggedright \end{minipage}}}%
    \put(0,0){\includegraphics[width=\unitlength,page=1]{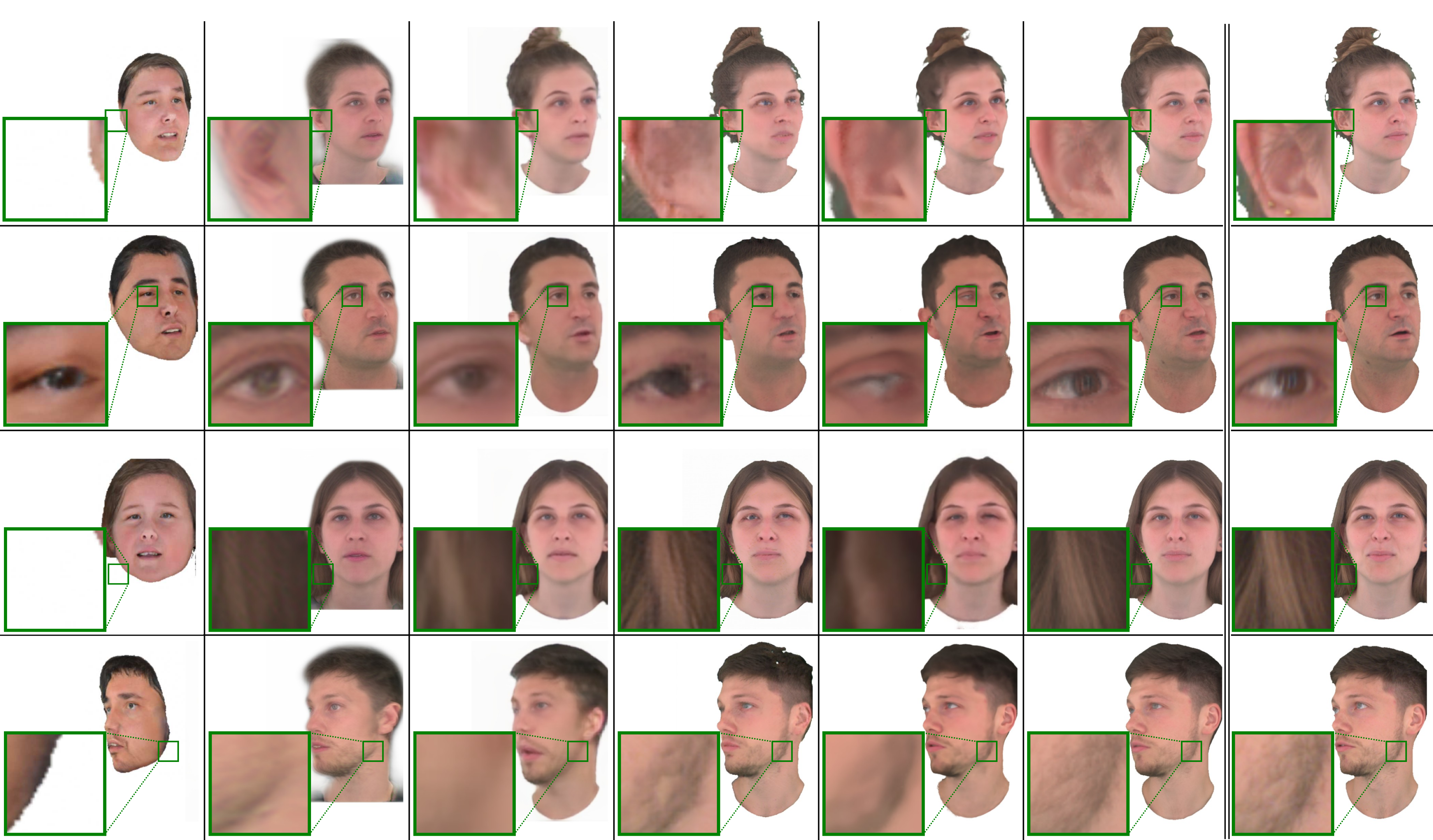}}%
  \end{picture}%
\endgroup%

    \caption{
    Comparison of novel pose \& expression synthesis results. VariTex, Bi-Layer and FOMM are one-shot approaches and estimate the avatar from the first frontal training frame. All other methods are optimized subject-specifically on the respective training set. %
    }
    \label{fig:qual_rec_eval}
\end{figure*}
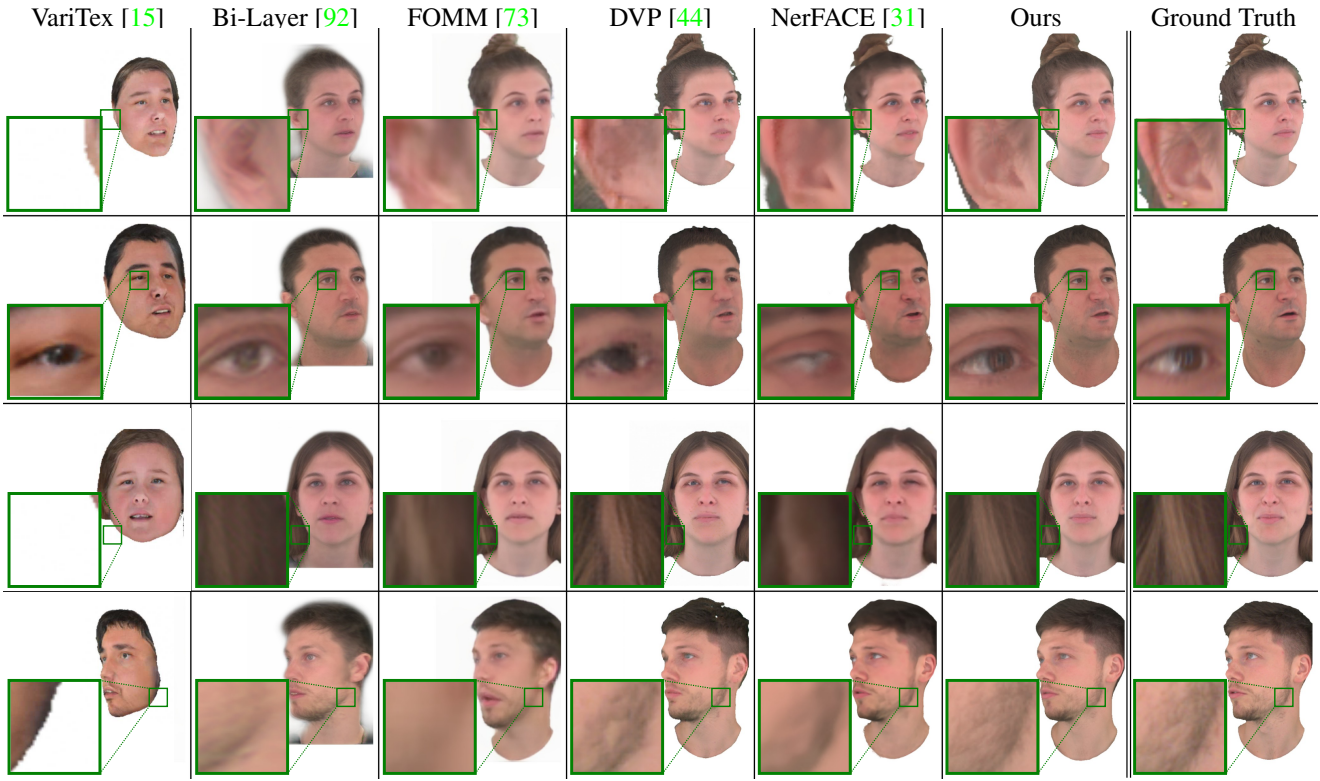

The animatable geometric backbone of our model allows to synthesize new expressions and poses for an optimized avatar.
We quantitatively and qualitatively evaluate our model on this task by optimizing it on the training partition of the real dataset and using the resulting avatar to reconstruct the validation frames.
For reconstructing the validation frames, we optimize the expression and pose parameters of our method in an analysis-by-synthesis manner, i.e., minimizing ~\Cref{eq:objective} only for $\psi$ and $\theta$,  while keeping all other components fixed.
We compare the results to recent works on talking-head synthesis:
NerFACE \cite{nerface}, Deep Video Portraits (DVP) \cite{deepvideoportraits}, First-Order Motion Model (FOMM) \cite{FOMM}, Bi-Layer \cite{bilayer}, and VariTex \cite{varitex}.
All one-shot approaches (FOMM, Bi-Layer and VariTex) are given the first frontal frame of the training sequence, while the subject-specific methods (NerFACE and DVP) utilize the whole training sequence.
For evaluation, we deploy the pixel-wise $\ell_1$ metric; the reference-free cumulative probability blur detection metric (CPBD) \cite{cpbd}; the multi-scale structural similarity metric (MS-SSIM)~\cite{ms_ssim}; the learned perceptual image patch similarity (LPIPS) \cite{lpips}; and the cosine distance of a pretrained face recognition network (CSIM) \cite{arcface}.
\Cref{tab:quant_rec_val} shows the resulting scores of perceptual and photometric metrics averaged over all sequences of our real dataset, and demonstrates that we outperform related approaches consistently. 
The qualitative comparison in \Cref{fig:qual_rec_eval} confirms that our model synthesizes images with higher detail and better expression conservation than related methods.
\begin{table}
    \resizebox{\linewidth}{!}{%
\begin{tabular}{l|ccccc} \toprule
\textbf{Method}& \textbf{L1} $\downarrow$& \textbf{MS-SSIM} $\uparrow$& \textbf{LPIPS} $\downarrow$& \textbf{CSIM} $\uparrow$& \textbf{CPBD} $\uparrow$\\ \midrule
VariTex \cite{varitex}& $0.207$& $0.640$& $0.316$& $0.347$& $0.565$\\
Bi-Layer \cite{bilayer}& $0.108$& $0.805$& $0.283$& $0.757$& $0.403$\\
FOMM \cite{FOMM}&$0.055$& $0.911$& $0.201$& $0.800$& $0.370$\\
DVP \cite{deepvideoportraits}& $0.042$& $0.904$& $0.098$& $0.815$& $0.632$\\
NerFACE \cite{nerface}& $0.057$& $0.897$& $0.156$& $0.864$& $0.485$\\
Ours& $\mathbf{0.033}$ & $\mathbf{0.923}$ & $\mathbf{0.079}$& $\mathbf{0.884}$& $\mathbf{0.674}$\\\bottomrule
\end{tabular}
}
\caption{Quantitative evaluation of the appearance reconstruction on the real dataset. Reported are the average scores over all validation frames for all subjects in our real dataset (see \Cref{sec:data}).}
\label{tab:quant_rec_val}
\end{table}

\subsection{Novel View Synthesis}
\begin{figure}
    \centering
    \includegraphics[width=\linewidth]{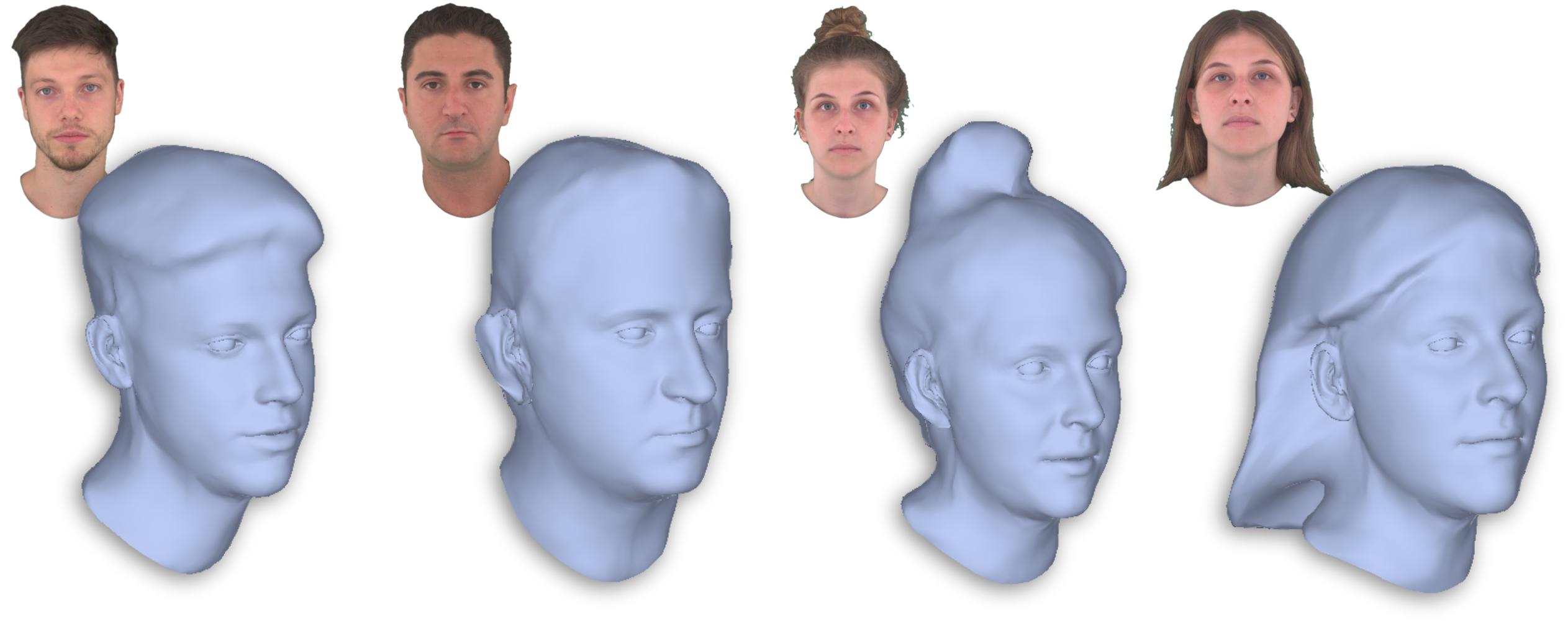}
    \caption{Our 3D mesh reconstruction for the subjects in the real dataset. The meshes align accurately with the real head shapes, even for longer hair, and preserve fine facial details.}
    \label{fig:geometry_real}
    \vspace{-4mm}
\end{figure}
The explicit geometry representation for each subject (shown in \Cref{fig:geometry_real}) enables 3D-consistent novel viewpoint synthesis.
We evaluate the novel viewpoint synthesis, by rendering frames from the validation set of the real sequences under novel yaw angles.
We compare our method against NerFACE \cite{nerface} and DeepVideoPortraits (DVP) \cite{deepvideoportraits} as these methods rely on a representation in 3D space that can be rotated accordingly.
The underlying 3D representations are centered on the image plane before rendering to account for varying coordinate origin definitions.
\Cref{fig:qual_novel_view_eval} demonstrates that while related methods suffer from significant artifacts, our method maintains its high visual quality.

\subsection{Ablation Study}

\begin{figure}[t]
    \centering
    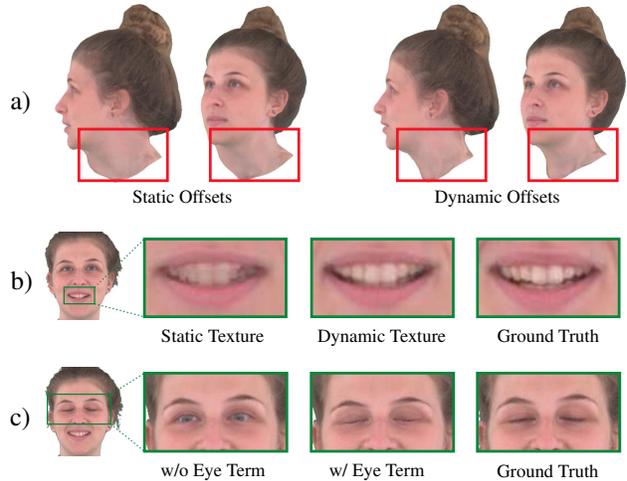
    \caption{Ablations: \textbf{a)} Static vertex offsets lead to bulky neck reconstructions compared to dynamic (pose-conditioned) geometry. \textbf{b)} Static texture representations fail to reconstruct highly dynamic face regions such as the interior of the mouth cavity. \textbf{c)} Explicit optimization for lid closing reconstructs eye closure faithfully.}
    \label{fig:ablations}
\end{figure}

We evaluate the influence of different architecture choices and optimization terms on the real dataset.
Our geometry is based on the FLAME model, and extends it to capture person-specific detail.
For deformations of the face, we rely on the blendshape-based expression model and the linear blend-skinning of the yaw bone.
Nevertheless, in our experiments, we found that a static geometry refinement of the underlying FLAME model does not reconstruct the neck region realistically as the joint poses and global rotation of the initial FLAME estimate are highly ambiguous.
To this end, we condition the geometry network $\mathcal{G}$ with the joint poses of the FLAME model which compensates errors in the neck region, see \Cref{fig:ablations}~(a).

Similar to the dynamic geometry network, we use a pose- and expression-dependent network $\mathcal{T}$ to predict the surface radiance.
\Cref{fig:ablations} (b) demonstrates that the use of a static texture, i.e. when the dynamic conditionings of $\mathcal{T}$ are fixed to zero, results in less authentic synthesis results, especially for the highly dynamic mouth region. 

Our model also supports eye blinks, which are modelled by the geometry of the FLAME model.
To enforce the reconstructions of eye blinks, we introduced a specific energy term on the landmarks in the eye region (see $E_\text{lmk}$ in \Cref{sec:method:optimization}).
Using this energy term in the shape optimization, the geometry faithfully reconstructs blinking eyes, see \Cref{fig:ablations} (c).
Without this term, eye blinks are not recovered.
\appx{Further energy term ablations are shown in the Appendix.}

\subsection{Discussion}

We have demonstrated that our method produces high quality results even for large head rotations due to our explicit geometry reconstruction of the face and hair region.
As such, our method addresses one of the main issues of other learning-based approaches which suffer from significant artifacts when synthesizing novel views.
Similar to all baselines, we do not address the synthesis of physical effects like floating or deforming hair.
Incorporating physics into a 4D avatar is an interesting field for future work.

\begin{figure}[t]
     \centering
     \begin{subfigure}[b]{0.15\textwidth}
         \centering
         \includegraphics[width=\textwidth]{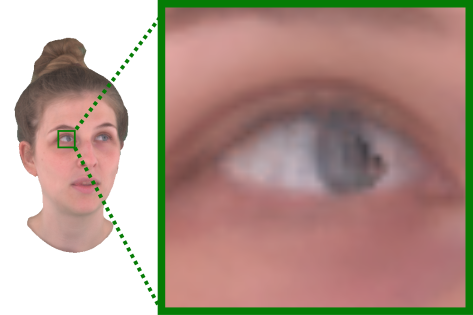}
         \caption{}
         \label{fig:y equals x}
     \end{subfigure}
     \hfill
     \begin{subfigure}[b]{0.15\textwidth}
         \centering
         \includegraphics[width=\textwidth]{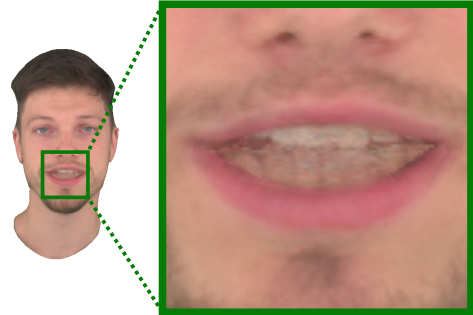}
         \caption{}
     \end{subfigure}
     \hfill
     \begin{subfigure}[b]{0.15\textwidth}
         \centering
         \def\svgwidth{\textwidth}
         \scriptsize{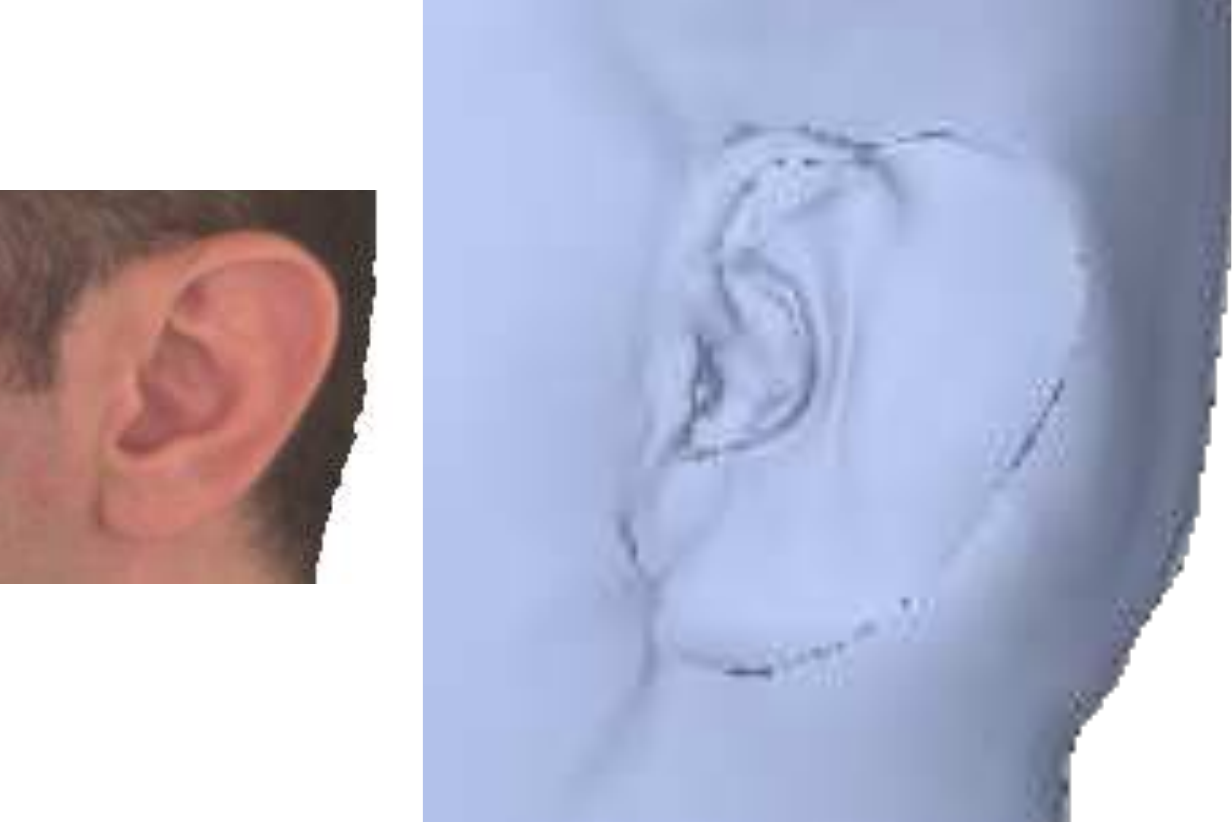}
         \caption{}
         \label{fig:five over x}
     \end{subfigure}
        \caption{Failure cases. Misaligned eye lids in the reconstructed mesh can cause rare eye artifacts (a). Extreme mouth poses may degrade the synthesized mouth interior (b). The reconstruction of ear shapes that deviate strongly from the statistical average yields local geometric artifacts (c).}
        \label{fig:failure_cases}
\end{figure}

Moreover, our method exhibits limitations in regions where the explicit geometry is greatly unconstrained, most prominently the mouth cavity.
As a consequence, the visual quality of the synthesized inner mouth region, especially teeth, may decrease if expressions and poses lie far outside of the training corpus (see \Cref{fig:failure_cases}).
The most natural approach to overcome this issue would be the integration of a well-aligned geometry of the inner mouth cavity.
However, so far none of the publicly available parametric head models provides such geometry due to the difficulties in the acquisition of ground truth data.
\begin{figure}[t]
    \centering
   \def\svgwidth{\linewidth}
   \small{\begingroup%
  \makeatletter%
  \providecommand\color[2][]{%
    \errmessage{(Inkscape) Color is used for the text in Inkscape, but the package 'color.sty' is not loaded}%
    \renewcommand\color[2][]{}%
  }%
  \providecommand\transparent[1]{%
    \errmessage{(Inkscape) Transparency is used (non-zero) for the text in Inkscape, but the package 'transparent.sty' is not loaded}%
    \renewcommand\transparent[1]{}%
  }%
  \providecommand\rotatebox[2]{#2}%
  \newcommand*\fsize{\dimexpr\f@size pt\relax}%
  \newcommand*\lineheight[1]{\fontsize{\fsize}{#1\fsize}\selectfont}%
  \ifx\svgwidth\undefined%
    \setlength{\unitlength}{1418.65806279bp}%
    \ifx\svgscale\undefined%
      \relax%
    \else%
      \setlength{\unitlength}{\unitlength * \real{\svgscale}}%
    \fi%
  \else%
    \setlength{\unitlength}{\svgwidth}%
  \fi%
  \global\let\svgwidth\undefined%
  \global\let\svgscale\undefined%
  \makeatother%
  \begin{picture}(1,0.8437551)%
    \lineheight{1}%
    \setlength\tabcolsep{0pt}%
    \put(-0.00301484,0.02213775){\color[rgb]{0,0,0}\rotatebox{90}{\makebox(0,0)[lt]{\begin{minipage}{0.81089382\unitlength}\centering \begin{tabularx}{\textwidth}{*3{>{\centering\arraybackslash}X}} Ours & NerFACE \cite{nerface} &  DVP \cite{deepvideoportraits}\\ \end{tabularx}\end{minipage}}}}%
    \put(0,0){\includegraphics[width=\unitlength,page=1]{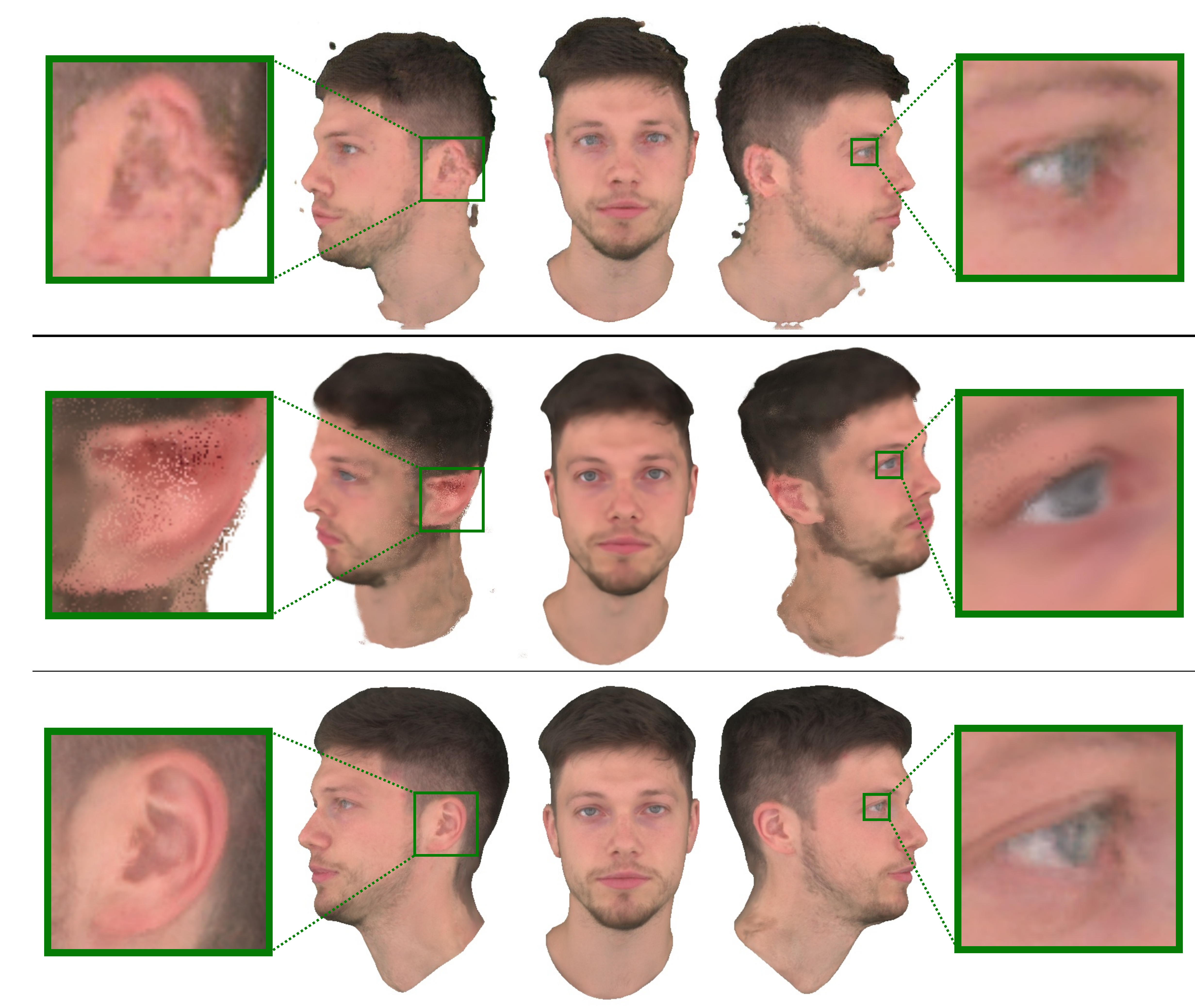}}%
  \end{picture}%
\endgroup%
}
    \caption[Qualitative Novel Viewpoint Synthesis Comparison]{Qualitative novel viewpoint synthesis comparison. While related methods suffer significant artifacts during novel viewpoint synthesis, our method exhibits high robustness even under large rotations and preserves its high texture detail.}
    \label{fig:qual_novel_view_eval}
\end{figure}
Similar to NerFACE~\cite{nerface} and DVP~\cite{deepvideoportraits}, our method is person-specific and, thus, requires optimization of the neural network for every new actor which takes $7$ hours using two Nvidia A100 GPUs, when optimizing on images with a resolution of $512\times512$ px. The optimization time and necessary computational resources can be reduced greatly when optimizing against smaller images or image patches.
Generalizing our approach is future work which can benefit from the findings of VariTex~\cite{varitex} and pi-GAN~\cite{piGAN}.

\paragraph{Ethical Considerations}
With the advances in the synthesis of photo-realistic human avatars, the potential misuse (e.g., misinformation) becomes an increasingly important ethical concern.
While active watermarking of generated content can be employed, there is no guarantee that this watermark can not be removed.
To this end, there exists the field of multi-media forensics which analyzes methods for active and passive forgery detection. %
Passive forgery detection~\cite{roessler2018faceforensics,roessler2019faceforensics++,cozzolino2018forensictransfer,cozzolino2020idreveal,fakeBehavior} is able to detect manipulated or synthetic imagery without any explicit watermarking.
While these methods can be trained to detect specific manipulation methods~\cite{roessler2018faceforensics,roessler2019faceforensics++}, generalized methods~\cite{cozzolino2018forensictransfer,cozzolino2020idreveal,fakeBehavior} have problems in reliably detecting fakes.
Also, forgery detectors can be used in an adversarial training to improve the quality of the synthetic renderings.
Thus, besides forgery detection algorithms, cryptographical approaches for signing the authenticity of video material have to be used in the future (which requires a trust network).

\section{Conclusion}
In this work, we presented Neural Head Avatars, a method that accurately reconstructs geometry and appearance of the human head from a monocular RGB sequence.
Our approach combines %
a parametric head model with multi-layer perceptrons that refine geometry and synthesize a photorealistic texture.
The resulting 4D avatar is robust with respect to large pose-, view- and expression changes, and we show that it outperforms state-of-the-art head avatar methods qualitatively and quantitatively. 

While recent work on head avatar synthesis moved towards implicit representations of spatial geometry, our work demonstrates the benefits of an explicit geometry reconstruction in combination with a deep appearance network for dynamic surface textures in terms of photorealism and generalizability.
We hope that our work inspires further research at the intersection of explicit geometry reconstruction and deep appearance representations. 

\section*{Acknowledgements}
This project has received funding from the DFG in the joint German-Japan-France grant agreement (RO 4804/3-1), the ERC Starting Grant Scan2CAD (804724), and a Microsoft Research Grant.
We thank the recorded actors and video narrator for their help, as well as, Mohit Mendiratta and Guy Gafni for providing the synthesis results of \cite{deepvideoportraits} and \cite{nerface} for our comparisons. 
We gratefully acknowledge the GWK support for funding this project by providing computing time through the Center for Information Services and HPC (ZIH) at TU Dresden.

{\small
\bibliographystyle{ieee_fullname}
\bibliography{egbib}
}

\clearpage

\appendix
\section{Implementation Details}
\label{sec:implementation_details}

\subsection{Template Model}
\begin{figure}
    \centering
    \begin{subfigure}[b]{0.49\linewidth}
         \centering
         \includegraphics[width=\linewidth]{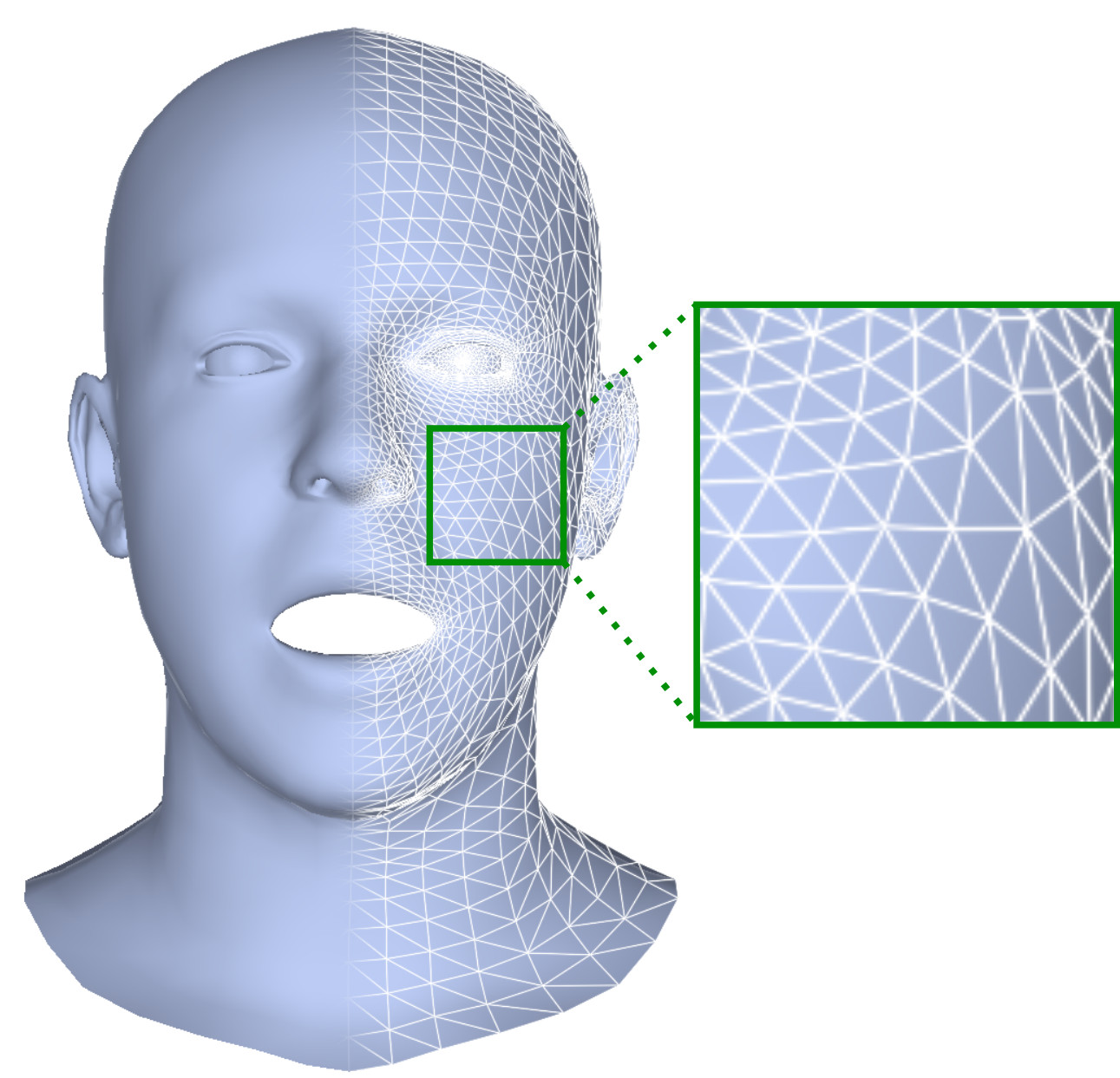}
         \caption{Original FLAME topology}
     \end{subfigure}%
     \begin{subfigure}[b]{0.49\linewidth}
         \centering
         \includegraphics[width=\linewidth]{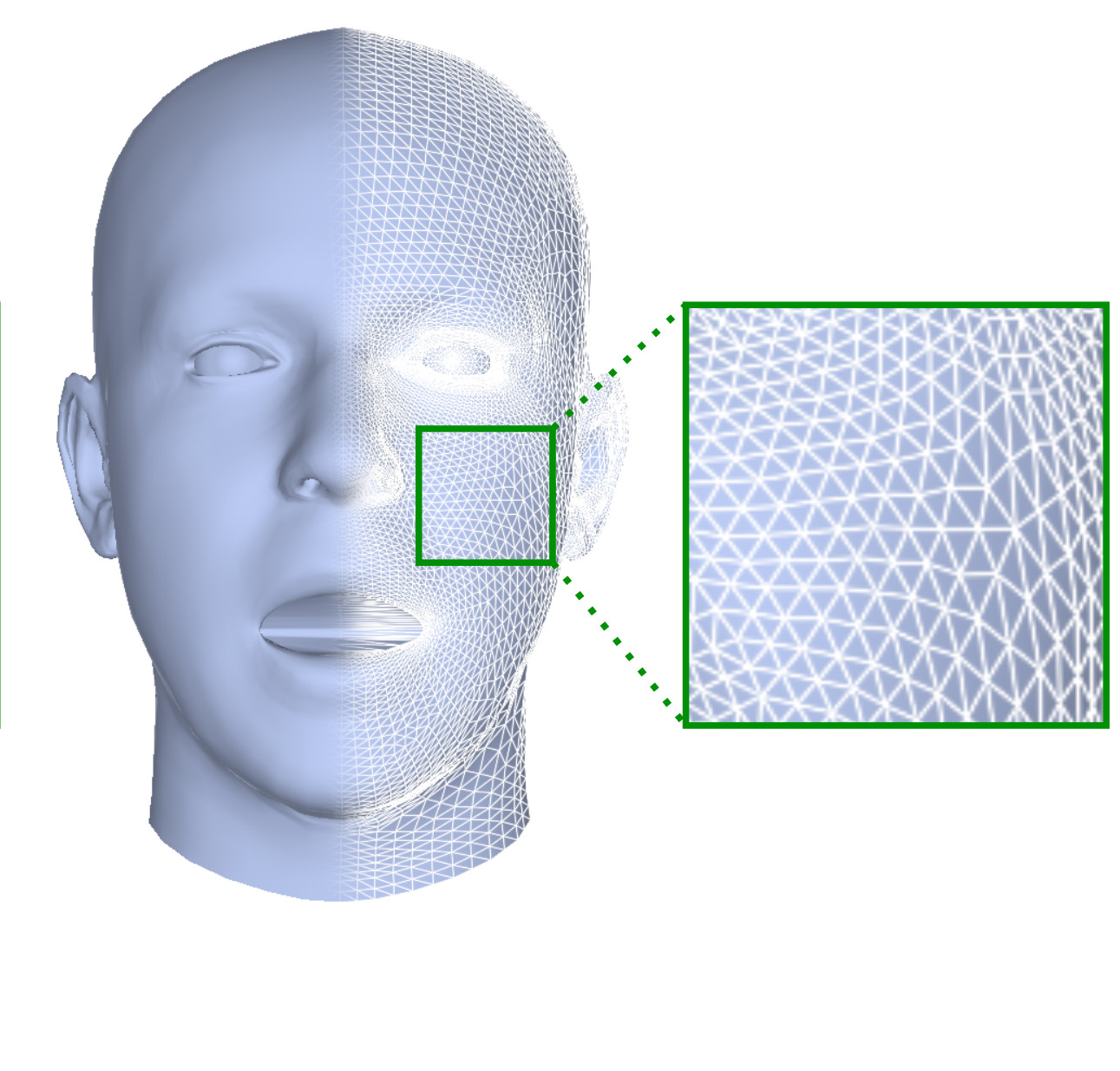}
         \caption{Our FLAME topology}
     \end{subfigure}
    \caption{As our template, we uniformly subdivide the FLAME~\cite{flame} mesh and simplify the mouth cavity.}
    \label{fig:flame}
\end{figure}

We utilize the FLAME head model \cite{flame} in its updated version from 2020 as the geometric backbone of our method.
As mentioned in the main paper, we perform minor adjustments to the FLAME topology, namely, we uniformly subdivide the faces (four-way subdivision), remove the faces belonging to the lower neck region, and add faces to close the mouth cavity, see \Cref{fig:flame}.
This increases the original number of vertices from 5023 to 16227.
Inspired by \cite{dynamicsurfacefunctions}, we use $\tanh$ activation functions to limit the joint rotations of FLAME to physiologically plausible ranges. 

\subsection{Network Architectures}
\label{appx:network-architectures}
\begin{figure*}
    \centering
    \includegraphics[width=\linewidth]{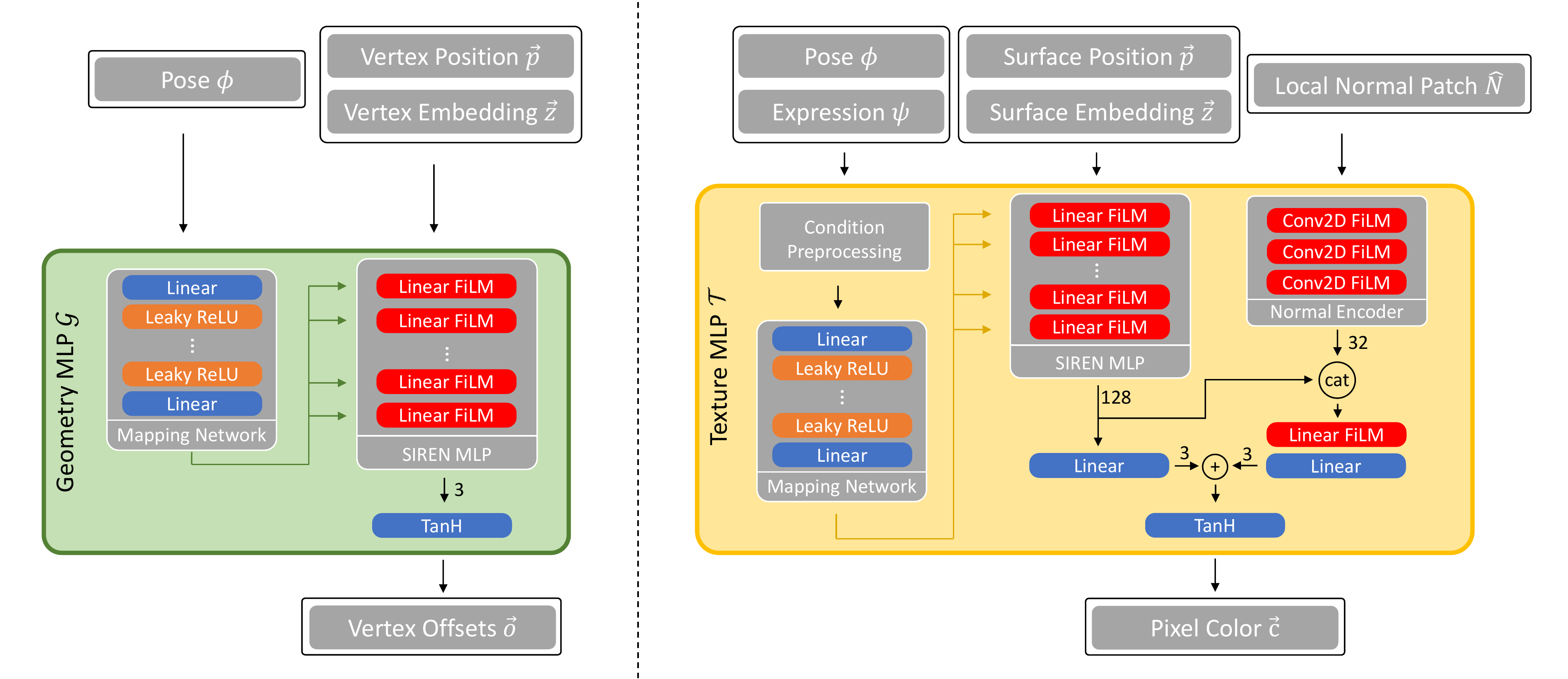}
    \caption{Overview of our model architectures. Our \textit{Neural Head Avatar} relies on SIREN-based MLPs \cite{siren} with fully connected linear layers, periodic activation functions and FiLM conditionings \cite{film, film2}. Inspired by \cite{piGAN}, surface coordinates and spatial embeddings (either vertex-wise for $\mathcal{G}$, or as an interpolatable grid in uv-space for $\mathcal{T}$) are used as an input to the SIREN MLP. The dynamic frequencies and phase shifts of the Linear FiLM layers are predicted by fully connected mapping networks which are conditioned on the FLAME parameters. For the Texture MLP $\mathcal{T}$, a fully convolutional normal encoder generates additional conditions from a local patch of the predicted normal map.}
    \label{fig:architecture}
\end{figure*}

Our method relies on two multi-layer perceptrons (MLPs). The Geometry MLP $\mathcal{G}$ refines the mesh resulting from the linear FLAME head model and adds facial detail and hair structure. The Texture MLP $\mathcal{T}$ synthesizes a dynamic, photo-realistic texture that is able to reproduce view- and expression dependent effects, e.g., wrinkles and reflections. This section details the model architecture of both networks. An overview is presented in \Cref{fig:architecture}.

\subsubsection{Geometry MLP \texorpdfstring{$\mathcal{G}$}{G}}
The Geometry MLP $\mathcal{G}$ adds facial detail and hair geometry to the mesh resulting from the FLAME head model. 
The inputs to the network are the 3D coordinates of the vertices on the template mesh (see \Cref{fig:flame}), normalized to the range $[-1, 1]$, as well as vertex-specific embedding vectors that are optimized during training. 
Dynamic geometry effects are enabled by also passing the FLAME pose parameters to $\mathcal{G}$. 
In practice, to avoid overfitting only the 3 pose parameters of the neck joint are used. 

The core of $\mathcal{G}$ is a SIREN-based fully connected MLP \cite{siren} which takes the vertex positions and embeddings as input, processes them by a sequence of FiLM-conditioned linear layers \cite{film, film2}, and predicts the three-dimensional vertex offsets. 
The linear, FiLM-conditioned layers first perform an affine transformation on the input signal, followed by a sinusoidal activation function. 
The phase shifts and frequencies of the sinusoidal activation functions are generated by a mapping network which is a concatenation of linear layers with leaky ReLU activations. This architecture was greatly inspired by \cite{piGAN}.
The input to the mapping network is the three-dimensional pose of the FLAME neck joint which enables the synthesis of dynamically changing geometry refinements. 
We ensure spatial geometry consistency by only allowing dynamic geometry refinements for the neck region. To this end, we compute the offsets twice: once conditioned on the orignial neck pose values and once with all conditions set to zero. We smoothly blend both offset predictions according to a fading body region mask. 

\paragraph{Configuration details.}
The vertex embedding vectors have a feature dimension of 32 (for each vertex of template mesh). The mapping network comprises 3 hidden layers with 256 neurons each, followed by leaky ReLU activation functions with slope 0.2. The SIREN MLP consists of 6 consecutive linear FiLM layers with 128 neurons each. The final layer is a linear layer with 3 output neurons, followed by a $\tanh$ activation function. 

\subsubsection{Texture MLP \texorpdfstring{$\mathcal{T}$}{G}}

The architecture of our texture representation $\mathcal{T}$ is similar to $\mathcal{G}$.
For obtaining the color value of a point on the mesh surface, its 3D coordinates on the FLAME template mesh as well as a surface embedding vector are fed into a SIREN MLP. The surface embedding vector is sampled from a discrete 2D grid in uv space. 
A mapping network takes features extracted from the FLAME pose and expression parameters and predicts the phase shifts and frequencies of the sinusoidal activation functions in the linear FiLM layers of the SIREN MLP to synthesize a dynamic texture.
The dynamic texture is needed in the regions where the reconstructed geometry does not align well with the target surface. 
This is especially the case for the mouth cavity. Consequently, we process the FLAME pose and expression parameters into features that are highly correlated with the mouth articulation. 
More specifically, we determine the effective rotation that applies to the mouth region, i.e., combine global rotation and neck rotation, and use its axis-angle representation together with 10 pair-wise distances between vertices on the inner side of the lips as inputs to the mapping network.
These conditions only cover mouth-related information and, therefore, are only used as input to the mapping network for the respective surface regions. For all other regions, we fill the conditioning vector with zeros. Effectively, this approach greatly limits the dynamic capacities of the $\mathcal{T}$ to ensure a close bound to the underlying geometry which in turn enables extrapolation to unseen poses and expressions.
The outputs of the mapping network are used to dynamically adjust the phase shifts and frequencies of the linear FiLM layers in the SIREN MLP. 
Given these conditions and the surface position and surface embedding, the SIREN MLP produces a latent space vector. 
This vector serves as input to two network heads. In the first head, the latent vector is fed through a linear layer with 3 output neurons. 
In the second head, the latent vector is concatenated with the outputs of a normal encoder network. This network takes a local patch of the rendered normal map as input, passes it through a sequence of 2D-convolutional FiLM layers and outputs a feature vector in latent space. 
2D-convolutional FiLM layers have the same structure as linear FiLM layers, but instead of a linear layer, a 2D-convolutional layer is used. 
The latent vector produced by the normal encoder contains information on the local geometry configuration and enables the synthesis of expression- and view-dependent effects (e.g., ambient occlusions and specular highlights). 
The output vectors of the SIREN MLP and the normal encoder are concatenated and fed through a sequence of linear layers to produce 3 output activations. 
The activations of both heads are summed up and a $\tanh$ activation is applied to achieve the final RGB predictions in a range of (-1, 1). 

\paragraph{Configuration details.}
The surface embedding is sampled in uv space from a discrete feature grid with $256 \times 256$ feature vectors with 64 channels each via bilinear interpolation. 
We use a separate uv map for the inner mouth region with $64 \times 64$ vectors. The mapping network has the same architecture as for the Geometry MLP $\mathcal{G}$. The SIREN MLP consists of 8 consecutive linear FiLM layers with 256 neurons each, except for the last which has 128 neurons. 
The normal encoder is designed as a fully convolutional network with 3 consecutive 2D-convolutional layers with stride 1 and kernel size 3 and with periodic activations. 
All layers have 128 feature channels except for the last one which has 32. The first model head contains one fully connected layer with 3 output neurons. The second model head contains one linear FiLM layer with 128 neurons and one linear layer with 3 neurons.

\begin{table*}[]
\centering
\begingroup
\renewcommand{\arraystretch}{1.5}
\begin{tabular}{l|cc|cc|cc|cc}
\multicolumn{1}{c|}{} &
  \multicolumn{2}{c|}{\Large $w_\text{reg,flame}$} &
  \multicolumn{2}{c|}{\Large $w_\text{reg,surface}$} &
  \multicolumn{2}{c|}{\Large $w_\text{reg,edge}$} &
  \multicolumn{2}{c}{\Large $w_\text{reg,lapl}$} \\
\multicolumn{1}{c|}{\textbf{}} &
  \renewcommand{\arraystretch}{1.}\begin{tabular}[c]{@{}c@{}}Geometry\\ Optim.\end{tabular} &
  \renewcommand{\arraystretch}{1.}\begin{tabular}[c]{@{}c@{}}Joint\\ Optim.\end{tabular} &
  \renewcommand{\arraystretch}{1.}\begin{tabular}[c]{@{}c@{}}Geometry\\ Optim.\end{tabular} &
  \renewcommand{\arraystretch}{1.}\begin{tabular}[c]{@{}c@{}}Joint\\ Optim.\end{tabular} &
  \renewcommand{\arraystretch}{1.}\begin{tabular}[c]{@{}c@{}}Geometry\\ Optim.\end{tabular} &
  \renewcommand{\arraystretch}{1.}\begin{tabular}[c]{@{}c@{}}Joint\\ Optim.\end{tabular} &
  \renewcommand{\arraystretch}{1.}\begin{tabular}[c]{@{}c@{}}Geometry\\ Optim.\end{tabular} &
  \renewcommand{\arraystretch}{1.}\begin{tabular}[c]{@{}c@{}}Joint\\ Optim.\end{tabular}\\ \hline
Eyeballs &
  1.0E-03 &
  1.0E-03 &
  1.0E-04 &
  1.0E-04 &
  0 &
  0 &
  0 &
  0 \\
Eye Surrounding &
  1.0E-03 &
  1.0E-03 &
  1.0E-04 &
  1.0E-04 &
  0 &
  0 &
  5 &
  10 \\
Forehead &
  1.0E-03 &
  1.0E-03 &
  1.0E-04 &
  1.0E-04 &
  0 &
  0 &
  0.05 &
  0.1 \\
Face &
  1.0E-03 &
  1.0E-03 &
  1.0E-04 &
  1.0E-04 &
  0 &
  0 &
  0.05 &
  0.1 \\
Ears &
  1.0E-03 &
  1.0E-03 &
  1.0E-04 &
  1.0E-04 &
  0 &
  0 &
  25 &
  50 \\
Scalp &
  1.0E-03 &
  1.0E-03 &
  1.0E-04 &
  1.0E-04 &
  10 &
  10 &
  0.05 &
  0.1 \\
Neck &
  1.0E-03 &
  1.0E-03 &
  1.0E-04 &
  1.0E-04 &
  0 &
  0 &
  0.1 &
  0.2 \\
Lower Neck &
  1.0E-03 &
  1.0E-03 &
  1.0E-04 &
  1.0E-04 &
  0 &
  0 &
  0.25 &
  0.5 \\
Nose &
  1.0E-03 &
  1.0E-03 &
  1.0E-04 &
  1.0E-04 &
  0 &
  0 &
  2.50E-02 &
  5.00E-02
\end{tabular}
\endgroup
\caption{Geometry regularization weights. The regularization weights differ for the individual head regions. However, they do not change among target subjects, i.e., they can be used to reconstruct a wide variety of geometries such as different hair styles.}
\label{tab:regweights}
\end{table*}

\subsection{Optimization from Monocular RGB Data}
\label{appx:optimization-from-monocular-rgb}
\paragraph{Detailed Geometry Objective $E_\text{geom}$ (Eq. 2)}
As defined in Eq. 2 in the main paper, the geometry energy term is:
\begin{equation}
    \begin{split}
    E_\text{geom} &=  w_\text{lmk} \cdot E_\text{lmk} + w_\text{normal} \cdot E_\text{normal}\\
                    &+w_\text{semantic} \cdot E_\text{semantic} +w_\text{reg,geom} \cdot E_\text{reg, geom}. \\
    \end{split}
\end{equation}
$E_\text{normal}$ and $E_\text{semantic}$ are already defined in the main paper, for the other terms we provide additional explanation.

The landmark energy $E_\text{lmk}$ measures the distance of detected 2D facial landmarks $\mathbf{l_i}\in \mathbb{R}^2$ and the projected counterparts on the mesh surface $\mathbf{\hat{l}_i}\in \mathbb{R}^2$ and is given by:
\begin{equation*}
    E_\text{lmk} = \sum_{i=1}^{70} ||\mathbf{l_i} - \mathbf{\hat{l}_i}||_1 + w_\text{lid} \cdot \sum_{i \in \{\text{left}, \text{right}\}} ||\mathbf{d_i} - \mathbf{\hat{d}_i}||_1.
\end{equation*}
Besides the absolute positions of the landmarks $\mathbf{\hat{l}_i}$, we measure relative distances $\mathbf{\hat{d}_i}$ of the eye landmarks at the upper and lower lid to improve the reconstruction of eye lid closure~\cite{deca}.
However, we found that target lid distances $\mathbf{d_i}$ are less noisy when being computed on the facial segmentation (computed by \cite{faceparsing}) rather than the detected landmarks.
The 2D facial landmarks $\mathbf{l_i}$ are detected using \cite{facealignment,mediapipe} and also contain two iris landmarks.

To avoid convergence to local minima, we employ several geometry regularization strategies summing up to the term $E_\text{reg, geom}$ which is given by:
\begin{equation}
    \begin{split}
        E_\text{reg,geom} &= w_\text{reg,flame} \cdot E_\text{reg,flame} + w_\text{reg,lapl} \cdot E_\text{reg,lapl}\\ &+ w_\text{reg,surface} \cdot E_\text{reg,surface} + w_\text{reg,edge} \cdot E_\text{reg,edge}.
    \end{split}
\end{equation}
Following \cite{3dmm,face2face}, $E_\text{reg,flame}$ uses the statistical properties of the linear shape model, and regularizes the prediction towards the average face:
\begin{equation*}
    E_\text{reg,flame} = w_\beta \cdot |\boldsymbol{\beta}|_2^2 + w_\theta \cdot |\boldsymbol{\theta}|_2^2 +  w_\psi \cdot |\boldsymbol{\psi}|_2^2.
\end{equation*}
The Laplacian regularizer $E_\text{reg,lapl}$ computes the relative loss of the Laplace values $\lambda^*(V)$ of the predicted mesh vertices $V$ w.r.t. the surface of the FLAME model $V_\text{Flame}$, and controls the smoothness of the predicted offsets:
\begin{equation*}
        E_\text{reg,lapl} = |\mathcal{W}_\text{reg,lapl} \circ \left( \lambda^*(V) - \lambda^*(V_\text{Flame}) \right)|_1 .
\label{eq:energies:normal}
\end{equation*}
Note that $\lambda^*(.)$ denotes the discretized Laplace-Beltrami operator on the 1-ring neighborhood.
$\mathcal{W}_\text{reg,lapl}\in \mathbb{R}_+^V$ defines vertex specific weights which allow to control the smoothness in specific regions (e.g., the neck region has a higher regularization).
$\circ$ denotes the component-wise Hadamard product.
As in \cite{dynamicsurfacefunctions}, we regularize pose-dependent offset variations by adding a pair-wise surface consistency term. For two randomly picked frames $i, j$, the $\ell_1$-norm is computed over the difference between $\mathcal{G}(\phi_i)$ and  $\mathcal{G}(\phi_j)$:
\begin{equation*}
        E_\text{reg,surface} =~ |\mathcal{G}(\phi_i) - \mathcal{G}(\phi_j)|_1.
\end{equation*}
\begin{figure*}[t]
     \centering
     \begin{subfigure}[b]{0.49\textwidth}
         \centering
         \def\svgwidth{\linewidth}
         {\scriptsize\begingroup%
  \makeatletter%
  \providecommand\color[2][]{%
    \errmessage{(Inkscape) Color is used for the text in Inkscape, but the package 'color.sty' is not loaded}%
    \renewcommand\color[2][]{}%
  }%
  \providecommand\transparent[1]{%
    \errmessage{(Inkscape) Transparency is used (non-zero) for the text in Inkscape, but the package 'transparent.sty' is not loaded}%
    \renewcommand\transparent[1]{}%
  }%
  \providecommand\rotatebox[2]{#2}%
  \newcommand*\fsize{\dimexpr\f@size pt\relax}%
  \newcommand*\lineheight[1]{\fontsize{\fsize}{#1\fsize}\selectfont}%
  \ifx\svgwidth\undefined%
    \setlength{\unitlength}{525.58126831bp}%
    \ifx\svgscale\undefined%
      \relax%
    \else%
      \setlength{\unitlength}{\unitlength * \real{\svgscale}}%
    \fi%
  \else%
    \setlength{\unitlength}{\svgwidth}%
  \fi%
  \global\let\svgwidth\undefined%
  \global\let\svgscale\undefined%
  \makeatother%
  \begin{picture}(1,0.71064176)%
    \lineheight{1}%
    \setlength\tabcolsep{0pt}%
    \put(-0.0081374,-0.00168584){\color[rgb]{0,0,0}\rotatebox{90}{\makebox(0,0)[lt]{\begin{minipage}{0.71232761\unitlength}\begin{tabularx}{\textwidth}{*3{>{\centering\arraybackslash}X}} Ours & NerFACE \cite{nerface} &  DVP \cite{deepvideoportraits}\\ \end{tabularx}\end{minipage}}}}%
    \put(0,0){\includegraphics[width=\unitlength,page=1]{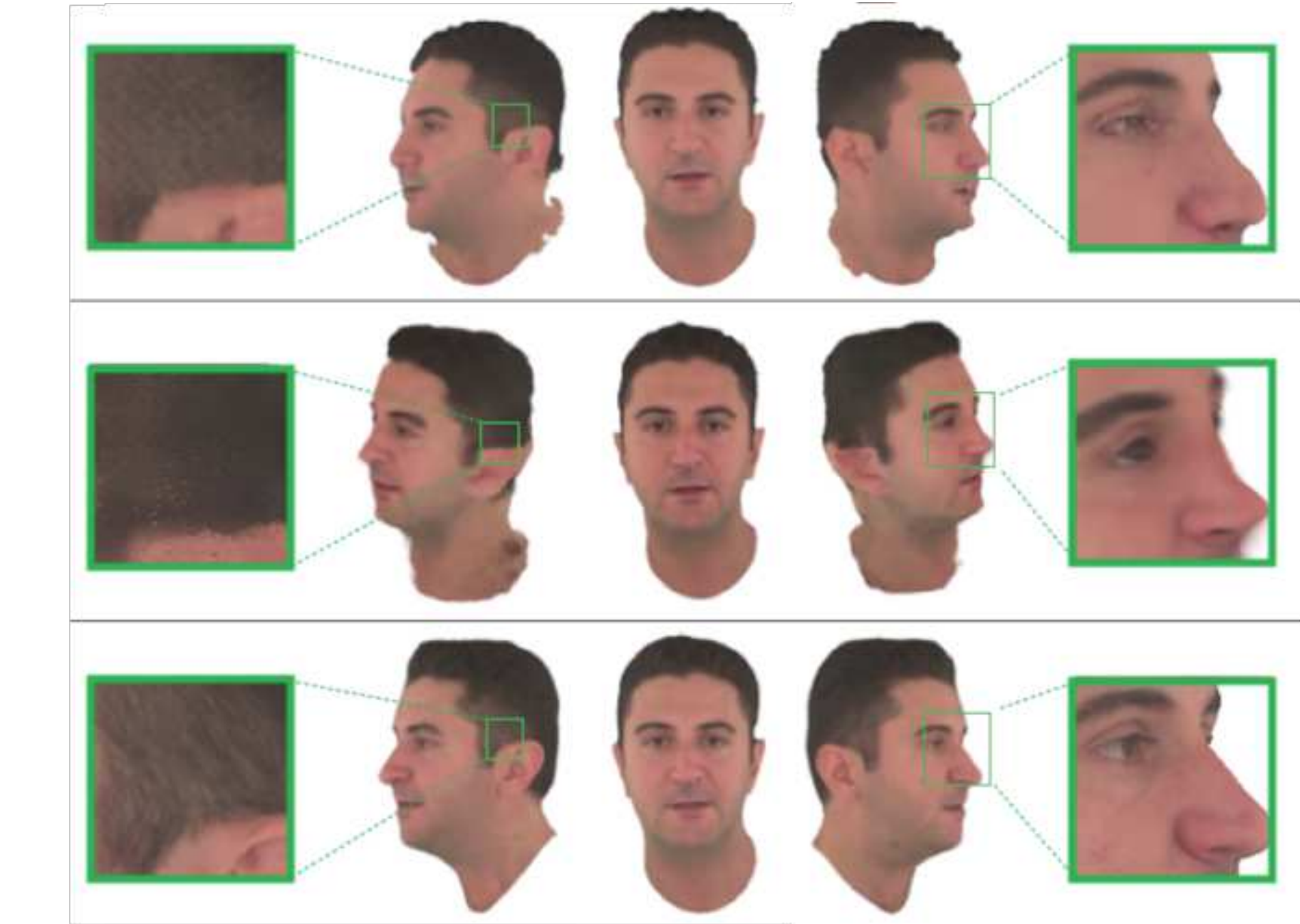}}%
  \end{picture}%
\endgroup%
}
         \label{fig:qual_novel_view_eval:radek}
     \end{subfigure}
     \hfill
     \begin{subfigure}[b]{0.49\textwidth}
         \centering
         \def\svgwidth{\linewidth}
         {\scriptsize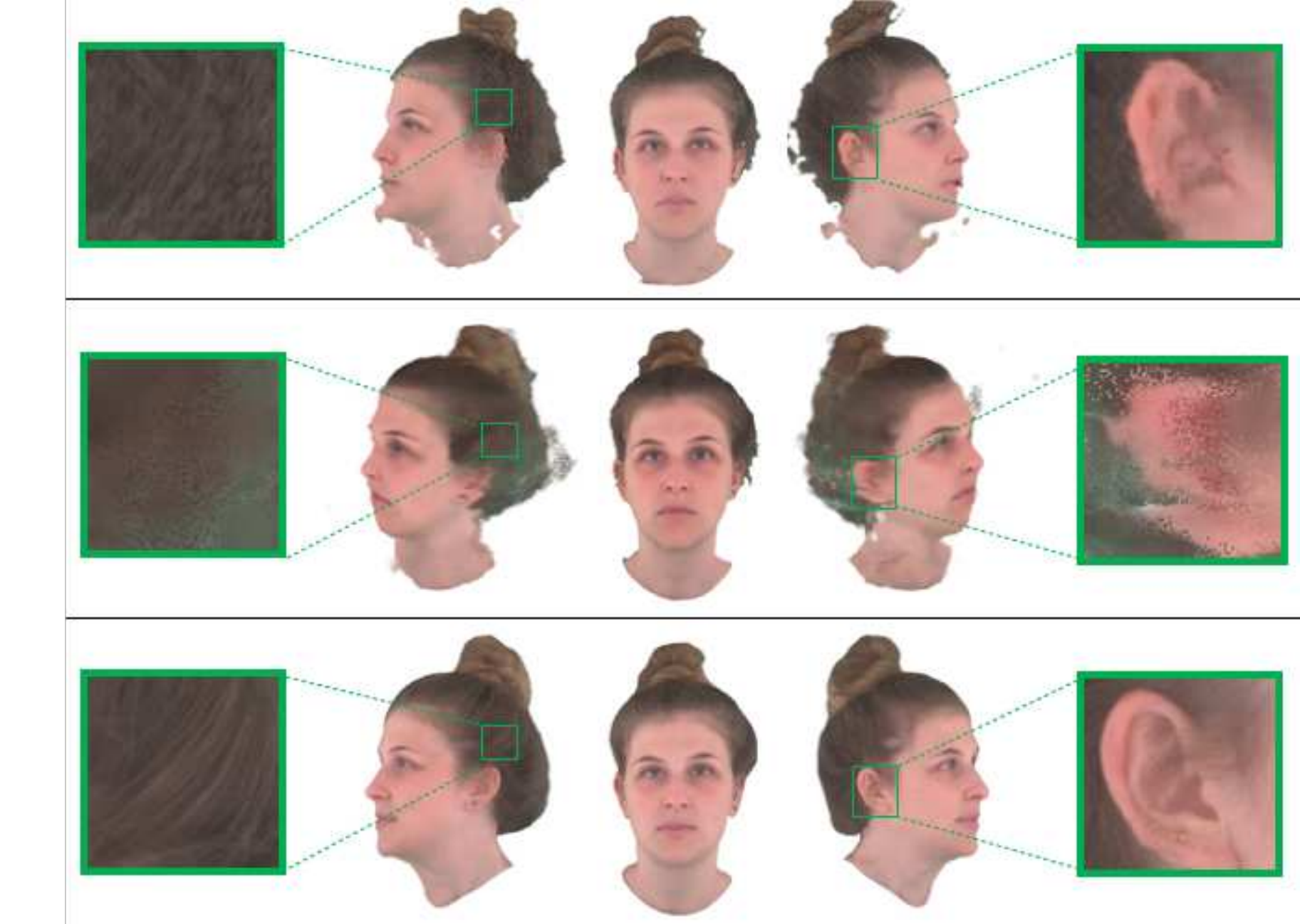}
         \label{fig:qual_novel_view_eval:julia}
     \end{subfigure}
     \vspace{-0.7cm}
    \caption{Additional qualitative novel view synthesis comparisons. We observe that the advantages of our method, namely spatial consistency and high texture detail even under extreme head rotations, apply to a variety of identities. Also see \Cref{fig:teaser} for additional subjects.}
    \label{fig:qual_novel_view_eval}
\end{figure*}

\begin{figure}[t]
     \centering
     \def\svgwidth{\linewidth}
     {\scriptsize\begingroup%
  \makeatletter%
  \providecommand\color[2][]{%
    \errmessage{(Inkscape) Color is used for the text in Inkscape, but the package 'color.sty' is not loaded}%
    \renewcommand\color[2][]{}%
  }%
  \providecommand\transparent[1]{%
    \errmessage{(Inkscape) Transparency is used (non-zero) for the text in Inkscape, but the package 'transparent.sty' is not loaded}%
    \renewcommand\transparent[1]{}%
  }%
  \providecommand\rotatebox[2]{#2}%
  \newcommand*\fsize{\dimexpr\f@size pt\relax}%
  \newcommand*\lineheight[1]{\fontsize{\fsize}{#1\fsize}\selectfont}%
  \ifx\svgwidth\undefined%
    \setlength{\unitlength}{242.92614331bp}%
    \ifx\svgscale\undefined%
      \relax%
    \else%
      \setlength{\unitlength}{\unitlength * \real{\svgscale}}%
    \fi%
  \else%
    \setlength{\unitlength}{\svgwidth}%
  \fi%
  \global\let\svgwidth\undefined%
  \global\let\svgscale\undefined%
  \makeatother%
  \begin{picture}(1,0.61468062)%
    \lineheight{1}%
    \setlength\tabcolsep{0pt}%
    \put(0,0){\includegraphics[width=\unitlength,page=1]{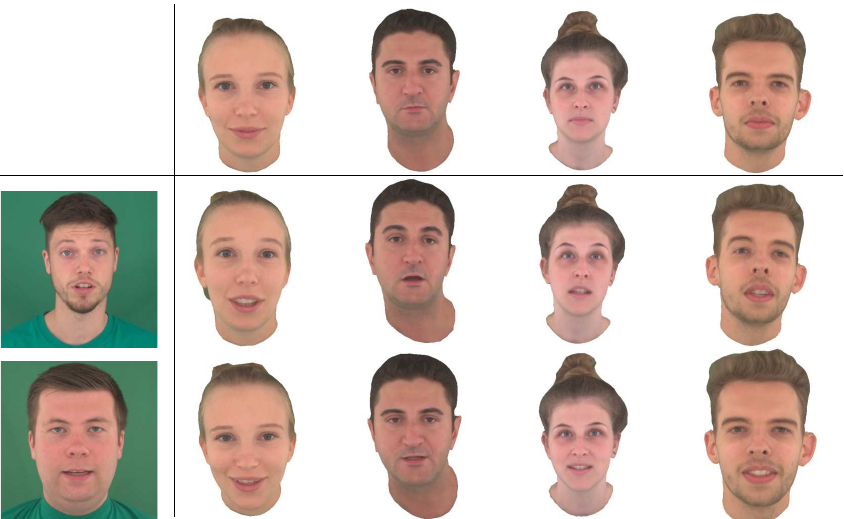}}%
    \put(0.05962685,0.41051193){\color[rgb]{0,0,0}\makebox(0,0)[lt]{\lineheight{1.25}\smash{\begin{tabular}[t]{l}Driver\end{tabular}}}}%
    \put(0.20168214,0.46156878){\color[rgb]{0,0,0}\rotatebox{90}{\makebox(0,0)[lt]{\lineheight{1.25}\smash{\begin{tabular}[t]{l}Source\end{tabular}}}}}%
  \end{picture}%
\endgroup%
}
     \caption{Expressions $\psi$ and poses $\phi$ from the driving frames on the left are added to the neutral poses of optimized source avatars at the top. The reenacted avatars are displayed below.}
     \label{fig:avatar_reenactment}
 \end{figure}

\begin{figure}[t]
     \centering
     \def\svgwidth{\linewidth}
     {\scriptsize\begingroup%
  \makeatletter%
  \providecommand\color[2][]{%
    \errmessage{(Inkscape) Color is used for the text in Inkscape, but the package 'color.sty' is not loaded}%
    \renewcommand\color[2][]{}%
  }%
  \providecommand\transparent[1]{%
    \errmessage{(Inkscape) Transparency is used (non-zero) for the text in Inkscape, but the package 'transparent.sty' is not loaded}%
    \renewcommand\transparent[1]{}%
  }%
  \providecommand\rotatebox[2]{#2}%
  \newcommand*\fsize{\dimexpr\f@size pt\relax}%
  \newcommand*\lineheight[1]{\fontsize{\fsize}{#1\fsize}\selectfont}%
  \ifx\svgwidth\undefined%
    \setlength{\unitlength}{1639.46887207bp}%
    \ifx\svgscale\undefined%
      \relax%
    \else%
      \setlength{\unitlength}{\unitlength * \real{\svgscale}}%
    \fi%
  \else%
    \setlength{\unitlength}{\svgwidth}%
  \fi%
  \global\let\svgwidth\undefined%
  \global\let\svgscale\undefined%
  \makeatother%
  \begin{picture}(1,0.7319012)%
    \lineheight{1}%
    \setlength\tabcolsep{0pt}%
    \put(0,0){\includegraphics[width=\unitlength,page=1]{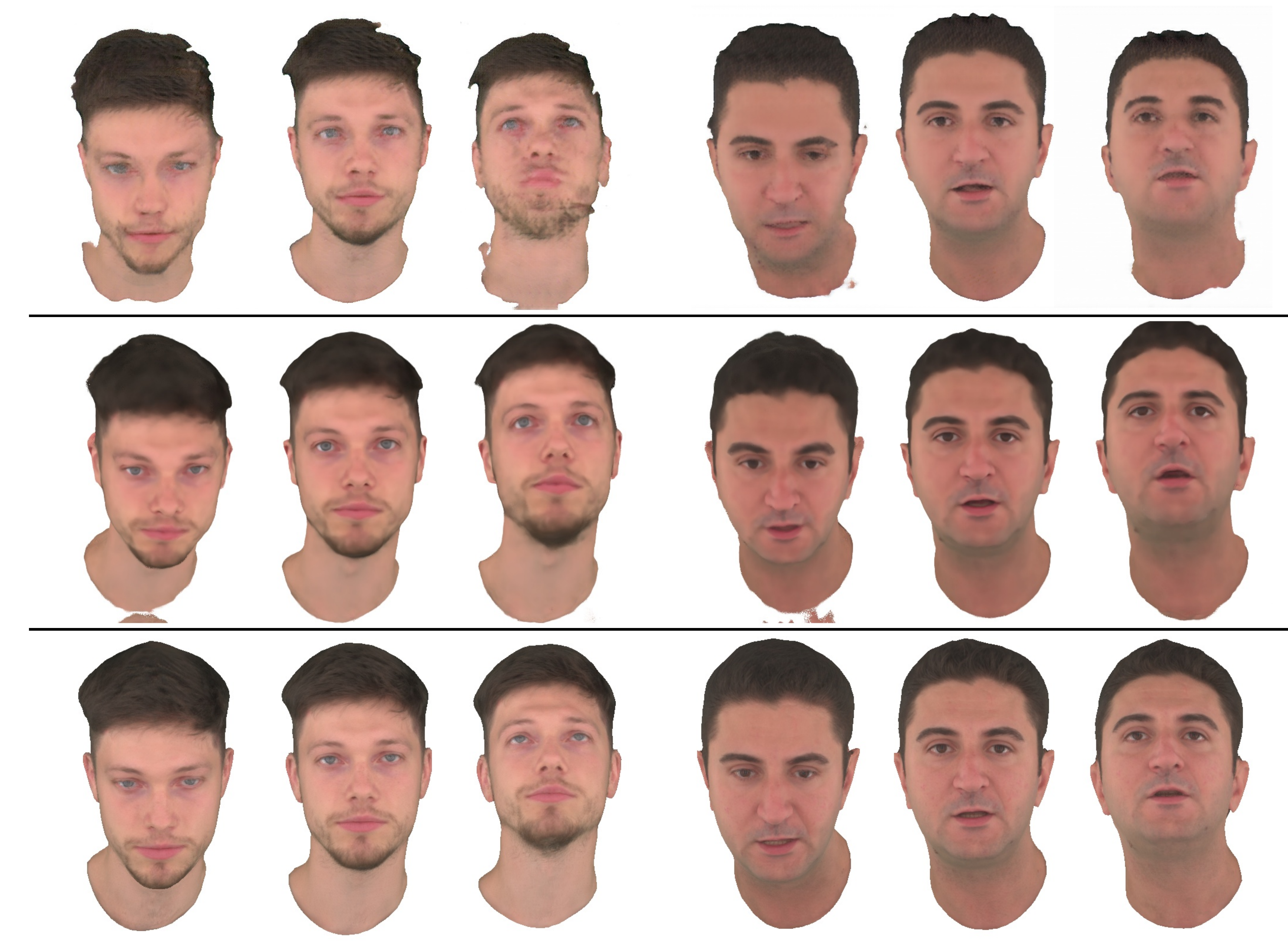}}%
    \put(-0.00260909,0.00024962){\color[rgb]{0,0,0}\rotatebox{90}{\makebox(0,0)[lt]{\begin{minipage}{0.72986493\unitlength}\begin{tabularx}{\textwidth}{*3{>{\centering\arraybackslash}X}} Ours & NerFACE \cite{nerface} &  DVP \cite{deepvideoportraits}\\ \end{tabularx}\end{minipage}}}}%
  \end{picture}%
\endgroup%
}
     \vspace{-0.7cm}
    \caption{Qualitative novel view synthesis comparison for pitch rotations. We report the frontal view as well as synthesis results under a pitch angle of $\pm 15$°.}
    \label{fig:qual_novel_view_eval_pitch}
\end{figure}
For large geometry corrections (e.g., long hair), we noticed an irregular vertex distribution over the surface which resulted in areas with large triangles and coarse shape modelling. To mitigate this issue, we regularize the length of edges $e_i$ in the scalp region if they deviate too much from the average edge length ${\bar{e}}$.
\begin{equation*}
        E_\text{reg,edge} =~ \sum_{e_i} \begin{cases}
                    e_i - \bar{e} & \text{if }  e_i > 1.5 \cdot \bar{e} \\
                     0 & \text{otherwise}
                 \end{cases}
\end{equation*}
As the individual facial regions are subject to different requirements, the applied regularization weights differ. \Cref{tab:regweights} presents the weights for the individual regions. Please note that while the weights differ for the individual face regions, they are the same for all avatars. No subject-specific fine-tuning of regularization parameters is required while still being able to reconstruct a wide set of structures (e.g. long and short hair). 

\paragraph{Detailed Appearance Objective $E_\text{app}$ (Eq. 3)}
The appearance energy term is defined in Eq. 3 in the main paper as:
\begin{equation}
    \begin{split}
        E_\text{app} &= w_\text{phot} \cdot E_\text{phot} + w_\text{perc} \cdot E_\text{perc} %
        .
    \end{split}
\end{equation}
The photo-metric energy term $E_\text{phot}$ is defined on the intersection $\mathcal{V} = \mathcal{S} \cap \mathcal{\hat{S}}$ of the foreground segmentation of the input $I$ and the region that is generated by our head model  $\hat{I}$: %
\begin{equation*}
    E_\text{phot} = | \mathcal{V} \cdot (\hat{I} - I) |_1 .
\end{equation*}
To generate sharp textures~\cite{larsen2016autoencoding,isola2017image}, we employ the style-based perceptual energy term $E_\text{perc}$ proposed by \cite{style_loss}.

\begin{figure*}[t]
     \centering
     \def\svgwidth{\linewidth}
         {\scriptsize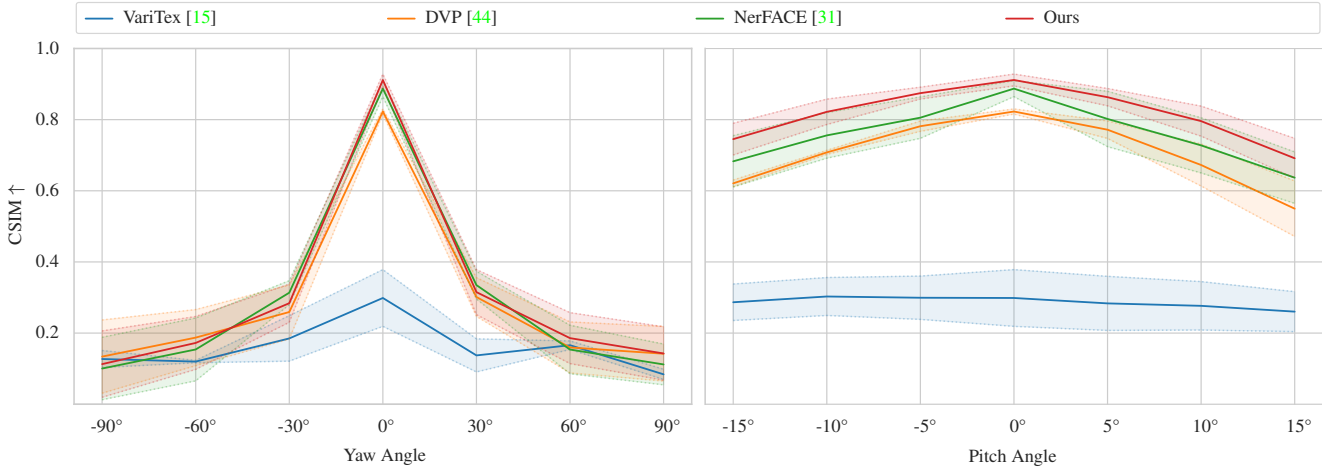}
    \caption{
    Quantitative novel view synthesis comparison. We report the cosine similarity score between latent features predicted by a pretrained face recognition network \cite{arcface}. The feature vectors are compared between the front facing ground truth and the predicted image under different rotation angles. We report the average scores over all validation frames from two sequences in our real dataset together with the respective $1\sigma$ regions. Our method consistently outperforms related approaches for pitch angles between -15° and +15°. For extreme yaw rotations, we observe significantly reduced CSIM scores even though qualitative comparisons demonstrate high identity preservation under these conditions (see Figure 7 in main paper).%
    }
    \label{fig:quant_novel_view_eval}
\end{figure*}
\begin{figure}[t]
     \centering
     \def\svgwidth{\linewidth}
     {\scriptsize\begingroup%
  \makeatletter%
  \providecommand\color[2][]{%
    \errmessage{(Inkscape) Color is used for the text in Inkscape, but the package 'color.sty' is not loaded}%
    \renewcommand\color[2][]{}%
  }%
  \providecommand\transparent[1]{%
    \errmessage{(Inkscape) Transparency is used (non-zero) for the text in Inkscape, but the package 'transparent.sty' is not loaded}%
    \renewcommand\transparent[1]{}%
  }%
  \providecommand\rotatebox[2]{#2}%
  \newcommand*\fsize{\dimexpr\f@size pt\relax}%
  \newcommand*\lineheight[1]{\fontsize{\fsize}{#1\fsize}\selectfont}%
  \ifx\svgwidth\undefined%
    \setlength{\unitlength}{244.07424996bp}%
    \ifx\svgscale\undefined%
      \relax%
    \else%
      \setlength{\unitlength}{\unitlength * \real{\svgscale}}%
    \fi%
  \else%
    \setlength{\unitlength}{\svgwidth}%
  \fi%
  \global\let\svgwidth\undefined%
  \global\let\svgscale\undefined%
  \makeatother%
  \begin{picture}(1,0.8539146)%
    \lineheight{1}%
    \setlength\tabcolsep{0pt}%
    \put(0,0){\includegraphics[width=\unitlength,page=1]{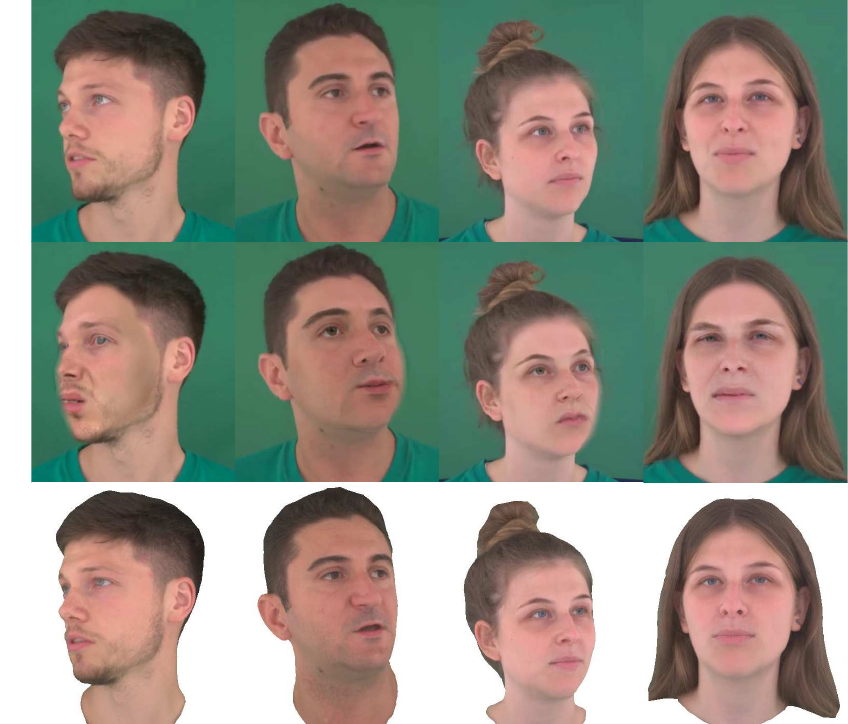}}%
    \put(0.01824499,0.09379839){\color[rgb]{0,0,0}\rotatebox{90}{\makebox(0,0)[lt]{\lineheight{1.25}\smash{\begin{tabular}[t]{l}Ours\end{tabular}}}}}%
    \put(0.01867711,0.36219857){\color[rgb]{0,0,0}\rotatebox{90}{\makebox(0,0)[lt]{\lineheight{1.25}\smash{\begin{tabular}[t]{l}paGAN \cite{pagan}\end{tabular}}}}}%
    \put(0.01824499,0.68402477){\color[rgb]{0,0,0}\rotatebox{90}{\makebox(0,0)[lt]{\lineheight{1.25}\smash{\begin{tabular}[t]{l}GT\end{tabular}}}}}%
  \end{picture}%
\endgroup%
}
    \caption{Qualitative comparison with paGAN \cite{pagan} on the validation sequences. Note that only the facial region is synthesized by paGAN.}
    \label{fig:pagan-comparison}
\end{figure}
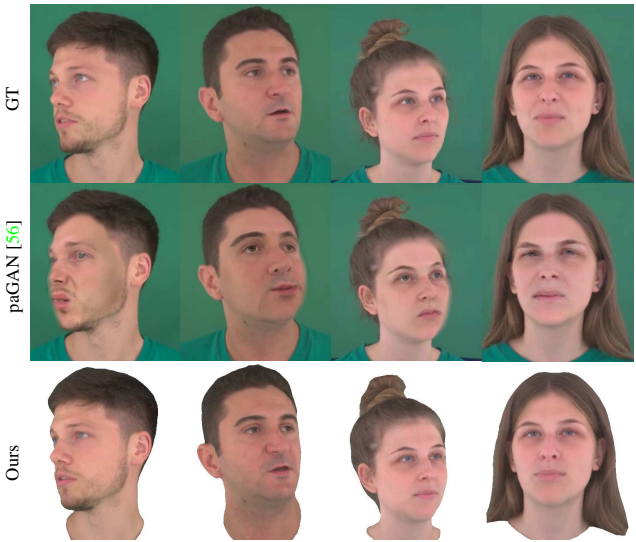
\paragraph{Initialization and Optimization.}
As discussed in Section 3.2 of the main paper, texture and geometry have to be initialized before starting the joint optimization against RGB sequences to avoid converging to bad local minima.
To this end, the FLAME head model is aligned with the training sequence to obtain a coarse geometry initialization.
Based on this coarse geometry, the weights of $\mathcal{G}$ and the frame-specific parameters of the FLAME model, i.e., pose and expressions are optimized w.r.t. $E_\text{geom}$ (Eq. (2) in main paper).
As a result of this optimization, we obtain a geometry estimate that aligns well with the target silhouette.
Using this refined geometry, $\mathcal{T}$ is trained to minimize $E_\text{app}$ (Eq. (3) in main paper).
In a final optimization step, all components are optimized jointly to minimize $E_\text{joint}$ (Eq. (1) in main paper).

Given the 750 input frames of our real sequences, we initialize $\mathcal{G}$ for 150 epochs, $\mathcal{T}$ is initialized for another 100 epochs, and we jointly optimize both components as long as the perceptual loss on the holdout validation set decreases.
This takes approximately 50 epochs.
For longer sequences, we reduce the epoch count such that the number of iterations per optimization stage remains the same.

We deploy a standard Adam optimizer \cite{adam} for all frame-agnostic parameters and a standard SGD optimizer for frame-specific components (e.g. expression and pose parameters).
Weight decay is applied to all texture-related components.
The entire pipeline is implemented with Pytorch \cite{pytorch} and Pytorch3D \cite{pytorch3d}.

\section{Additional Results}

\subsection{Avatar Reenactment}
\emph{Neural Head Avatars} are controllable by the disentangled pose and expression spaces of the FLAME head model.
This naturally enables the reenactment of an optimized avatar via a driving sequence.
We demonstrate this capability in \Cref{fig:avatar_reenactment}.
In this experiment, we utilize expressions $\psi$ and poses $\phi$ from a driver's video and add them to the neutral expression of optimized avatars.
$\psi$ and $\phi$ are extracted using our tracking algorithm discussed in the main paper.
We observe that our method is able to faithfully transfer pose and expression between various subjects.

\subsection{Novel View Synthesis}
In Section 4.4 of the main paper, we qualitatively compare the synthesis from novel viewpoints against related methods.
\Cref{fig:qual_novel_view_eval_pitch} and \Cref{fig:qual_novel_view_eval} provide further qualitative comparisons on different subjects and demonstrate that the advantages of our method apply to a variety of identities. 

We also provide a quantitative evaluation of the novel view synthesis results in \Cref{fig:quant_novel_view_eval}.
For a quantitative analysis, given that no ground truth is available for novel views, we evaluate the cosine similarity (CSIM) of the latent feature vectors predicted by a pretrained face recognition network \cite{facealignment} between front-facing ground truth and novel view prediction. 
We report the CSIM scores averaged over all validation frames of two sequences in our real dataset.
We find that our method outperforms related approaches consistently for pitch angles in a range of $\pm 15$°. 
However, we observe that for large yaw angles ($\ge\pm 30$°), the scores for all considered models decrease rapidly.  Still, qualitative comparisons demonstrate that our method exhibits high identity preservation even under viewpoint changes in that range.

\subsection{Comparison with paGAN\cite{pagan}}

In \Cref{fig:pagan-comparison}, we compare our generated images of the real identities with the outputs of paGAN\cite{pagan} provided by the authors. As their results are overlays on top of the original video, only the synthesized parts (facial region) should be considered when comparing to it.

\subsection{Synthesis Results for Non-Caucasian Subjects}

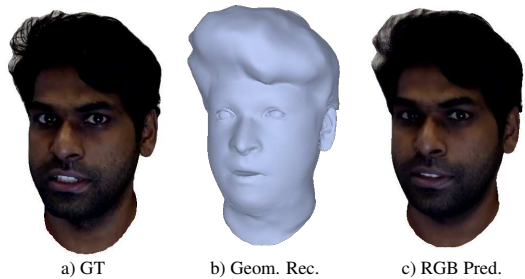
\begin{figure}
     \centering
     \def\svgwidth{\linewidth}
     {\scriptsize\begingroup%
  \makeatletter%
  \providecommand\color[2][]{%
    \errmessage{(Inkscape) Color is used for the text in Inkscape, but the package 'color.sty' is not loaded}%
    \renewcommand\color[2][]{}%
  }%
  \providecommand\transparent[1]{%
    \errmessage{(Inkscape) Transparency is used (non-zero) for the text in Inkscape, but the package 'transparent.sty' is not loaded}%
    \renewcommand\transparent[1]{}%
  }%
  \providecommand\rotatebox[2]{#2}%
  \newcommand*\fsize{\dimexpr\f@size pt\relax}%
  \newcommand*\lineheight[1]{\fontsize{\fsize}{#1\fsize}\selectfont}%
  \ifx\svgwidth\undefined%
    \setlength{\unitlength}{909.02929111bp}%
    \ifx\svgscale\undefined%
      \relax%
    \else%
      \setlength{\unitlength}{\unitlength * \real{\svgscale}}%
    \fi%
  \else%
    \setlength{\unitlength}{\svgwidth}%
  \fi%
  \global\let\svgwidth\undefined%
  \global\let\svgscale\undefined%
  \makeatother%
  \begin{picture}(1,0.47217539)%
    \lineheight{1}%
    \setlength\tabcolsep{0pt}%
    \put(0,0){\includegraphics[width=\unitlength,page=1]{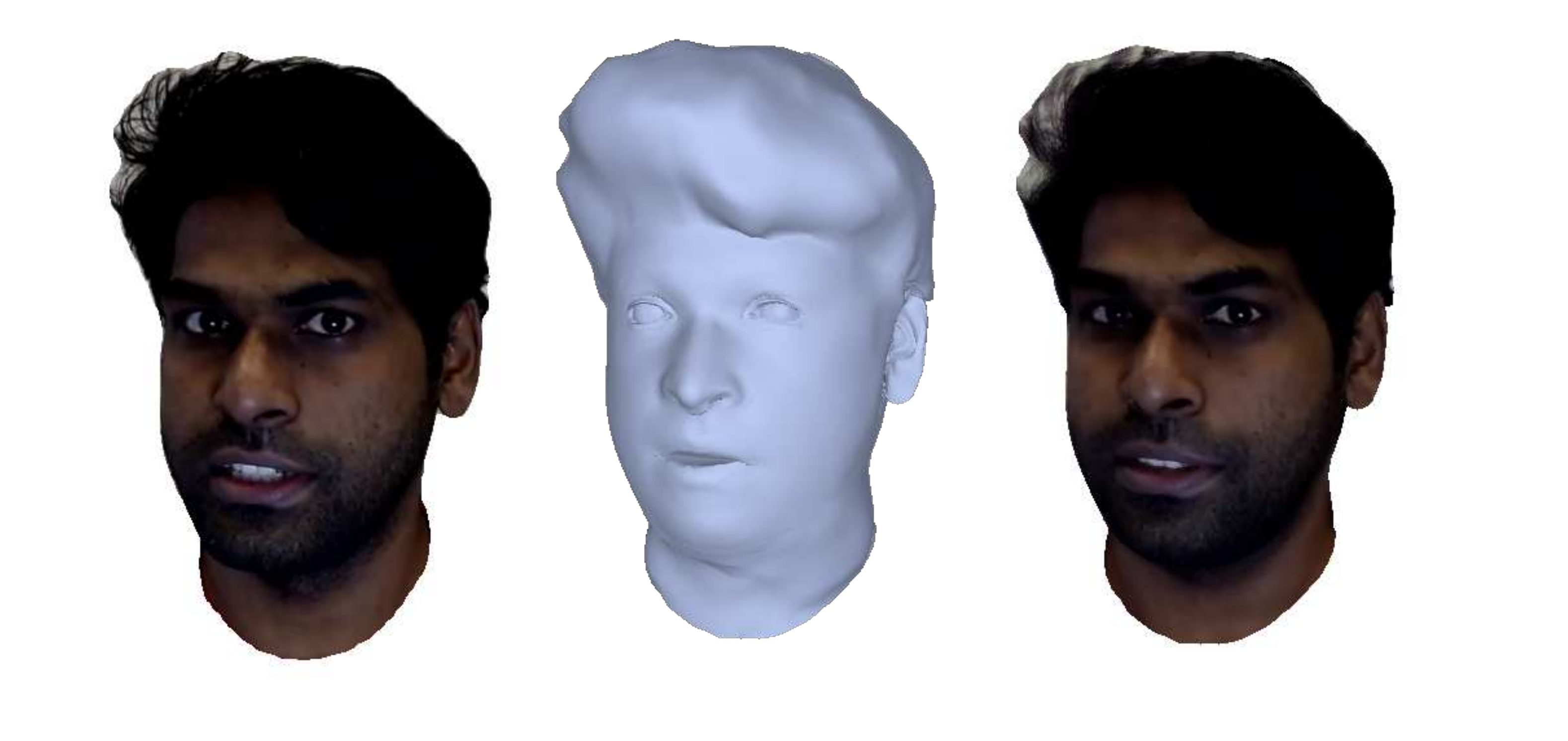}}%
    \put(0.1885011,0.01921486){\makebox(0,0)[t]{\lineheight{1.25}\smash{\begin{tabular}[t]{c}a) GT\end{tabular}}}}%
    \put(0.47842782,0.01921486){\makebox(0,0)[t]{\lineheight{1.25}\smash{\begin{tabular}[t]{c}b) Geom. Rec.\end{tabular}}}}%
    \put(0.7767559,0.01921486){\makebox(0,0)[t]{\lineheight{1.25}\smash{\begin{tabular}[t]{c}c) RGB Pred.\end{tabular}}}}%
  \end{picture}%
\endgroup%
}
     \caption{Synthesis results for a non-caucasian subject. The high quality of the synthesis results of our method is consistent also for people of color.}
     \label{fig:bala}
\end{figure}

To validate that our method also performs well for non-caucasian subjects, we include synthesis results for a person of color in \Cref{fig:bala}. Also in this case, our method reconstructs the head geometry faithfully and renders visually plausible images. 

\subsection{Geometry Evaluation}
\begin{figure}
    \centering
    \includegraphics[width=\linewidth,trim={1cm 0 0.5cm 0},clip]{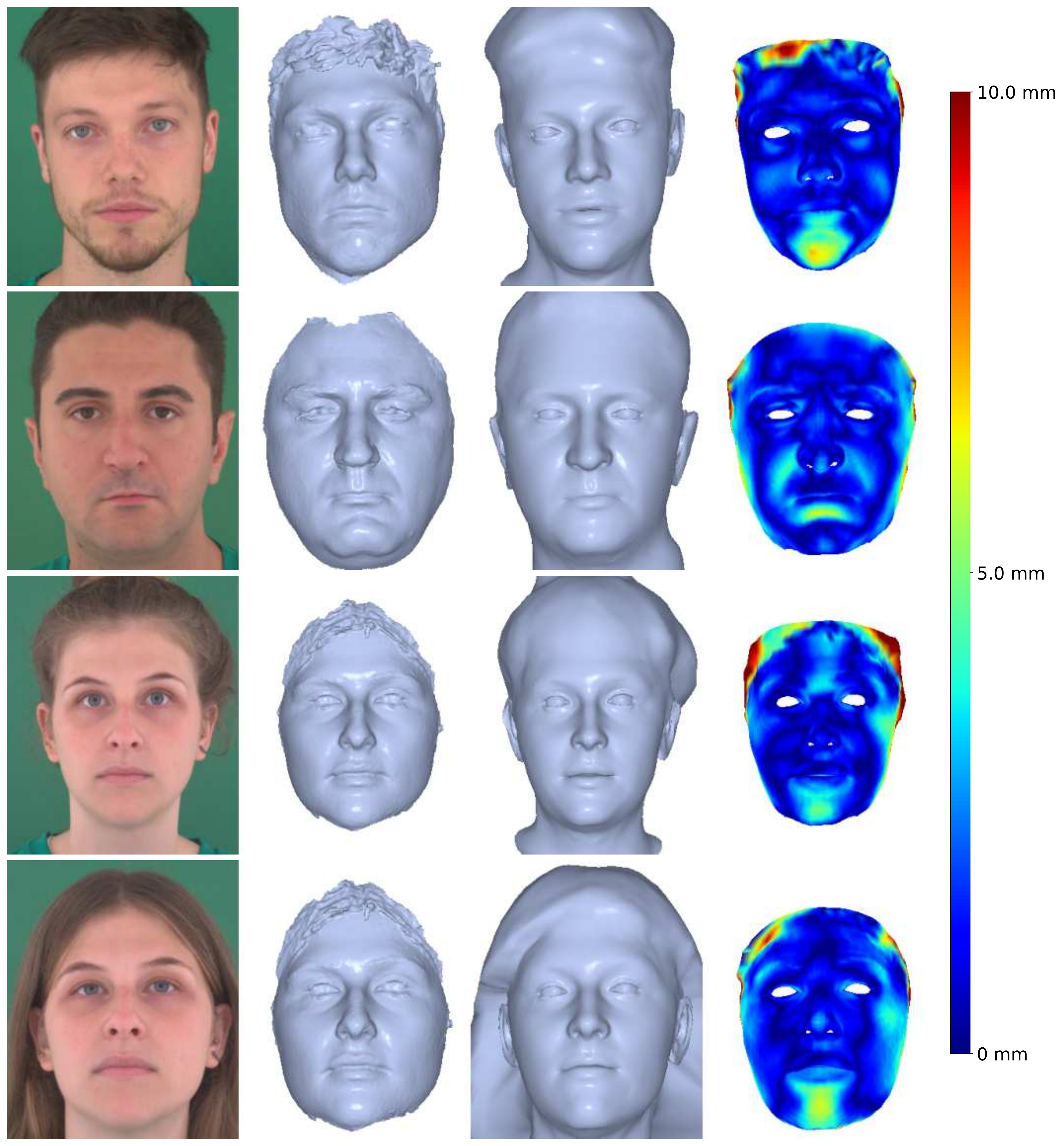}
    \caption[Mesh-to-mesh comparison with real subjects]{Geometry evaluation on the real identities. The right most column visualizes the Hausdorff distance from our predicted face mesh to the recorded GT. From top to bottom, the total alignment errors of the identities are $1.5$, $1.6$, $1.6$, and $1.6$ mm.}
    \label{fig:geometry-evaluation-real}
\end{figure}

We compare the geometry predicted by our approach on the real dataset with multi-view stereo (MVS) recordings of the respective identities, see Figure~\ref{fig:geometry-evaluation-real}. The MVS data was captured separately with a handheld DSLR camera. As hair styles and face dynamics can not be reproduced reliably in separate recordings, only the neutral pose of the face region is compared.

\subsection{Energy Ablation}
\begin{figure}[h]
    \centering
    \def\svgwidth{\linewidth}
    {\footnotesize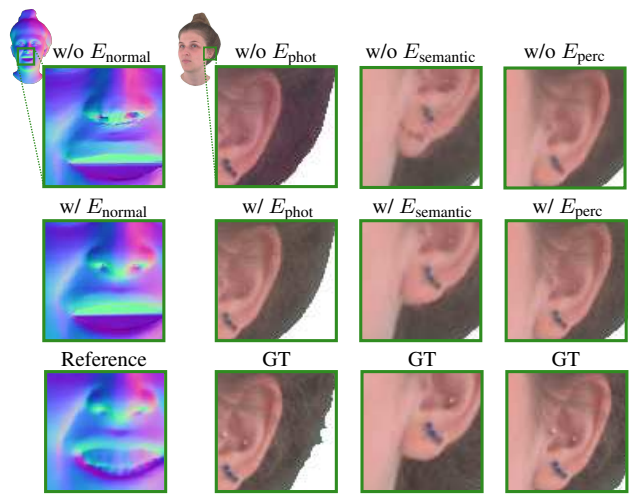}
    \caption{Energy Term Ablation. Reference normals are the inputs we use for optimization.}
    \label{fig:energy_ablation}
\end{figure}

\Cref{fig:energy_ablation} demonstrates the effect of further energy terms on the synthesis results. We observe that $E_\text{phot}$ prevents color shifts, $E_\text{semantic}$ improves alignment of overlapping semantic regions within the avatar's silhouette and $E_\text{perc}$ results in textures with more detail.

\end{document}